\documentclass[lettersize,journal]{IEEEtran}
\usepackage{amsmath,amsfonts}
\usepackage{algorithmic}
\usepackage{algorithm}
\usepackage{array}
\usepackage[caption=false,font=normalsize,labelfont=sf,textfont=sf]{subfig}
\usepackage{textcomp}
\usepackage{stfloats}
\usepackage{url}
\usepackage{verbatim}
\usepackage{graphicx}
\usepackage{cite}
\usepackage[utf8]{inputenc}
\usepackage{amsthm}
\usepackage{amssymb}
\usepackage{bm}
\usepackage{bbm}
\usepackage{soul}
\usepackage{booktabs}
\usepackage{multirow}
\usepackage{makecell}
\usepackage{tabularx}
\usepackage[table,xcdraw]{xcolor}
\usepackage{arydshln} % 必须放在 xcolor 之后
\usepackage{pifont}
\usepackage{circledsteps}
\usepackage{adjustbox} 
\usepackage{enumitem} 

\usepackage[hidelinks]{hyperref} 
\usepackage[hidelinks]{hyperref}
\hypersetup{
	colorlinks=true,       % 开启文字颜色，自动关闭方框
	linkcolor=blue,        % 内部引用颜色 (如 Figure 1, Table 2, Equation 3)
	filecolor=blue,        % 本地文件链接颜色
	urlcolor=blue,         % 网页 URL 颜色 (如 github 链接)
	citecolor=blue         % 参考文献引用颜色 (如 [1], [2])
}
\newcommand{\citeye}[1]{\textit{{\footnotesize #1}}}
\newcommand{\mc}[1]{\multicolumn{3}{c|}{#1}}

\newcommand{\cnum}[1]{{\footnotesize\Circled{#1}\normalsize}}
\newcommand{\myparagraph}[1]{
	\vspace{1.5ex}
	\noindent\textbf{#1}\hspace*{5pt}
}

\begin{document}

\title{Dual-Stream Semantic Guidance with Prototype Anchor Calibration for Source-Fully-Free Adaptation of Vision-Language Models}

\author{Weiwei~Xiang,
	Shun~Peng,
	Guangyi~Xiao$^*$,~\IEEEmembership{Member,~IEEE,}
	Hao~Chen,
	and~Lei~Yang$^*$,% <-this % stops a space
%	\thanks{This work was supported in part by the National Natural Science Foundation of China under Grants 62372167 and 62272156, and in part by the Natural Science Foundation of Hunan Province of China under Grants 2025JJ70455 and 2024JJ5097.}% <-this % stops a space
	\thanks{W. Xiang, S. Peng, G. Xiao, H. Chen, and L. Yang are with the College of Computer Science and Electronic Engineering, Hunan University, Changsha 410082, China (e-mail: mr\_menand@hnu.edu.cn; pengshun@hnu.edu.cn; gyxiao@hnu.edu.cn; chenhao@hnu.edu.cn; jt\_yl@hnu.edu.cn).}% <-this % stops a space
	\thanks{W. Xiang is also with Huaihua University, Huaihua 418008, China.}% <-this % stops a space
	\thanks{Corresponding authors: Guangyi Xiao and Lei Yang.}% <-this % stops a space
}

\markboth{IEEE Transactions on Image Processing}
{Xiang \MakeLowercase{\textit{et al.}}: DSSG: Dual-Stream Semantic Guidance}

%\IEEEpubid{0000--0000/00\$00.00~\copyright~2021 IEEE}
% Remember, if you use this you must call \IEEEpubidadjcol in the second
% column for its text to clear the IEEEpubid mark.

\maketitle

\begin{abstract}
Source-Fully-Free Domain Adaptation (SFF-DA) has emerged as a strategic paradigm to adapt Vision-Language Models (VLMs) without any access to source data or task-specific source models. However, we identify a critical \textit{Dual Semantic Drift} that hinders this process: static drift arising from the rigidity of fixed class embeddings, and dynamic drift stemming from the divergence of generated captions, causing severe semantic misalignment that intensifies the stability-plasticity dilemma.

To address this, we propose DSSG (Dual-Stream Semantic Guidance), an end-to-end framework that reconciles fine-grained plasticity with global stability. Our core contribution is the Dual Semantic Guidance (DSG) module, which integrates a caption stream for domain-specific knowledge with a class-anchor stream to anchor global categorical consistency. Furthermore, a Dynamic Cross-Modal Knowledge Distillation (CMKD) module is introduced to leverage the evolving teacher distribution for calibrating teacher-student consistency. Building upon DSSG, we further introduce Prototype Anchor Calibration (PAC), yielding DSSG-PAC, which periodically calibrates prototype anchors and caches them until the next calibration. This design reduces redundant text-side computation while preserving the adaptability of class guidance to the evolving text space. We further establish SFF-DA risk bounds that relate student risk to semantic-teacher quality and teacher--student discrepancy. Extensive experiments demonstrate that DSSG consistently outperforms current state-of-the-art methods across multiple benchmarks, while DSSG-PAC largely preserves its adaptation performance with 18.9\% lower total adaptation time. The code is available at \href{https://github.com/mrmenand/DSSG}{https://github.com/mrmenand/DSSG}.

\end{abstract}

\begin{IEEEkeywords}
Source-Fully-Free Domain Adaptation, Vision-Language Models, Cross-Modal Knowledge Distillation 
\end{IEEEkeywords}

\section{Introduction}

\IEEEPARstart{T}{raditional} unimodal source-free domain adaptation (SFDA) follows a two-stage pipeline of source-domain pre-training and target-domain adaptation. While effective, this paradigm heavily relies on task-specific source priors and is restricted by practical deployment bottlenecks, such as data privacy concerns and model black-box constraints (e.g., API-only access). Recently, Vision-Language Models (VLMs) represented by CLIP, pre-trained on large-scale image-text pairs, exhibit strong zero-shot generalization, serving as generalized source models rich in transferable knowledge. To eliminate thhe dependency on task-specific source training, we focus on Source-Fully-Free Domain Adaptation (SFF-DA)—also known as Unsupervised Fine-Tuning (UFT)~\cite{li2026clip}. 
We directly leverage VLMs for adaptation on unlabeled target domains without accessing any source data or task-specific source models, aiming to transfer their inherent universal knowledge to domain-specific contexts under distribution shifts.

\begin{figure}[t]    
	\centering
	\begin{adjustbox}{width=0.99\linewidth,center}
		\includegraphics{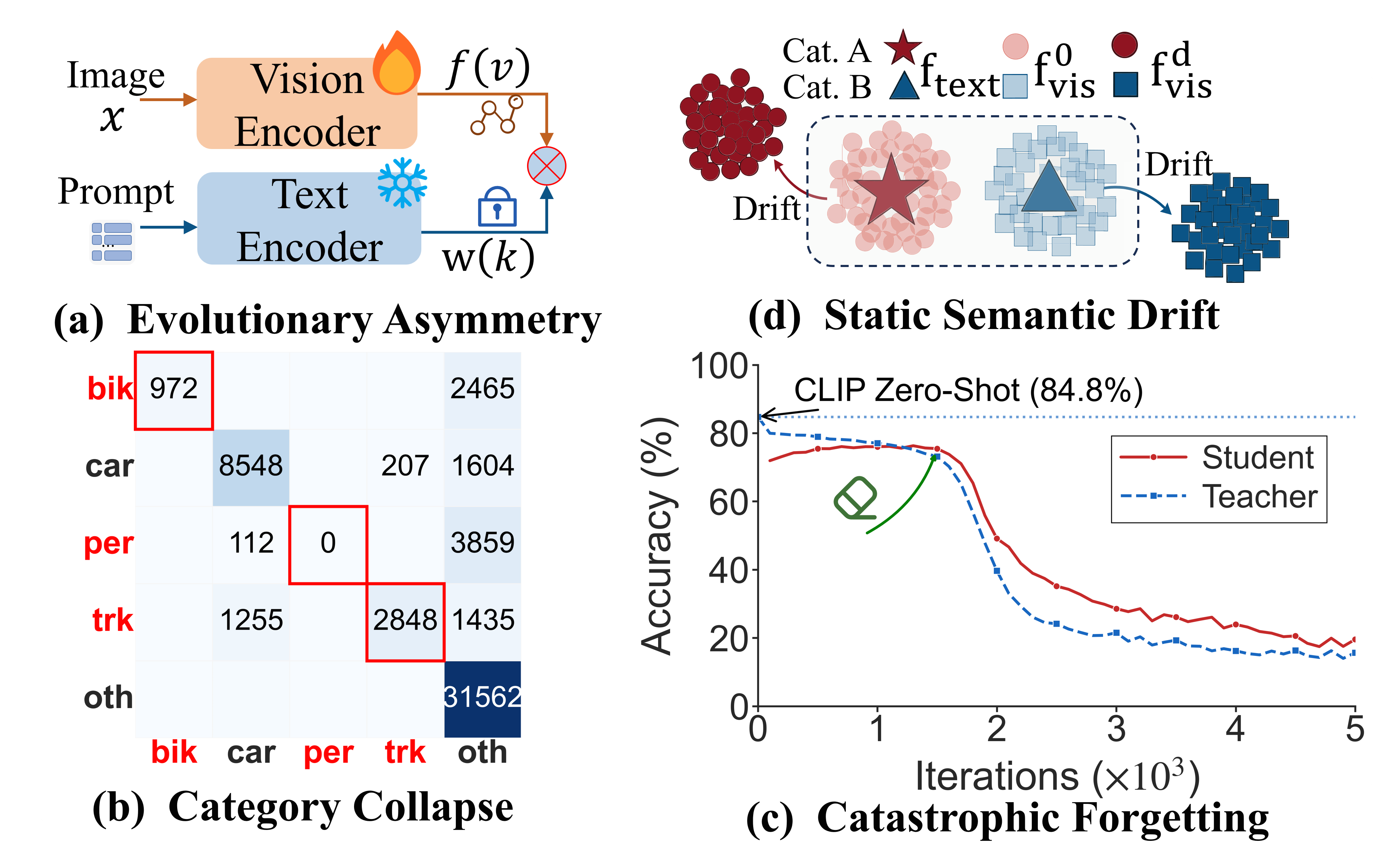}
	\end{adjustbox} 
\caption{Challenges of directly fine-tuning CLIP in SFF-DA. It can induce (b) Category Collapse and (c) Catastrophic Forgetting, while visual adaptation with fixed text embeddings causes (a) Evolutionary Asymmetry and (d) Static Semantic Drift. Here, $0/d$ denote the initial and drifted visual features, respectively.}
\label{Fig-challenge}

\end{figure}

While parameter-efficient methods (e.g., Prompt Learning~\cite{dapl2023}, Adapters~\cite{zhang2022tip}) preserve VLM priors by freezing backbones, their limited learnable parameters hinder the acquisition of intricate domain-specific knowledge. In contrast, directly fine-tuning foundation models in SFF-DA encounters a twofold challenge: \textit{(i) error accumulation and catastrophic forgetting} (Fig.~\ref{Fig-challenge}(b,c)), where the lack of source guidance accumulates pseudo-label errors, exacerbating confirmation bias and category collapse, thereby corrupting large-scale pre-trained knowledge; and \textit{(ii) evolutionary asymmetry and static semantic drift} (Fig.~\ref{Fig-challenge}(a, d)), where the vision encoder evolves toward a domain-specific manifold while template-derived embeddings $w_k$ remain fixed in the frozen text space, inducing a spatiotemporal misalignment that undermines cross-modal consistency, termed static semantic drift.

Existing CLIP-based SFDA methods generally follow two paradigms. First, CLIP-assisted approaches (e.g., DIFO and ProDe~\cite{difo2024,25prode}) employ CLIP as an external teacher to calibrate task-specific source models. However, their reliance on pre-trained source weights limits their applicability to SFF-DA. Second, unsupervised fine-tuning approaches adapt VLMs by optimizing prompt variables~\cite{24tfupt}, fine-tuning the vision encoder~\cite{23lafter}, or jointly tuning the vision and text encoders~\cite{24reclip}. Although encoder updates improve target-domain adaptability, they may also compromise the generalizable knowledge acquired during pre-training. Together, these limitations expose a fundamental stability--plasticity dilemma: prompt-based tuning provides insufficient plasticity to capture target-specific distributions, whereas encoder-based tuning risks damaging pre-trained knowledge.

\myparagraph{Motivation:} 
Drawing inspiration from ImCapDA~\cite{25imcapda}, we leverage VLM-generated captions as instance-level textual guidance and align them with target images through contrastive learning, enabling the vision and text encoders to jointly acquire target-domain semantics while facilitating cross-modal alignment. However, the absence of source supervision in the SFF-DA setting leaves the open-vocabulary semantics introduced by generated captions unconstrained, making caption-guided adaptation prone to dynamic semantic drift. Specifically, captions may emphasize non-discriminative instance details, background or out-of-distribution semantics, or cues from other task classes, reducing their reliability as category-discriminative guidance. Meanwhile, the fixed text class embeddings $w_k$ remain anchored in the original text space and cannot track the evolving visual feature space, resulting in static semantic drift. Together, caption-induced dynamic semantic drift and static semantic drift from fixed class embeddings constitute the \textit{Dual Semantic Drift} dilemma, making it difficult to acquire domain-specific knowledge while preserving global categorical consistency.

To address this, we present the Dual-Stream Semantic Guidance (DSSG) framework. First, the DSG module integrates a caption stream and a class-anchor stream to provide complementary semantic guidance: captions supply instance-specific context to mitigate forgetting, while class anchors preserve intrinsic category semantics. Second, the Dynamic CMKD module leverages the evolving teacher distribution through self-distillation to enforce teacher--student consistency. Together, these modules promote cross-modal alignment while preserving global categorical consistency. Building on DSSG, we introduce Prototype Anchor Calibration (PAC), yielding DSSG-PAC, which replaces iteration-wise class-prompt encoding with periodic prototype calibration and caching to reduce text-side training overhead while preserving adaptive category guidance.

Our main contributions are summarized as follows:
 \begin{enumerate} 
 	\item[1.] We identify Dual Semantic Drift in SFF-DA, arising from static class embeddings and unconstrained caption semantics, and establish risk bounds that relate student risk to semantic-teacher quality and teacher--student discrepancy.
	 \item[2.] We propose DSSG, which integrates dual-stream semantic guidance with Dynamic CMKD to enable caption-guided adaptation while preserving global categorical consistency.
 	\item[3.] We extend DSSG with Prototype Anchor Calibration, yielding DSSG-PAC, which replaces iteration-wise class-prompt encoding with periodic prototype calibration to reduce training overhead.
 	\item[4.] Extensive experiments demonstrate that DSSG achieves new state-of-the-art performance across multiple benchmarks, while DSSG-PAC largely preserves its accuracy with 18.9\% lower total adaptation time.
	\end{enumerate}

\section{Related Work}

\myparagraph{Unimodal Source-Free Domain Adaptation.} Unimodal SFDA methods typically rely solely on visual representations and leverage task-specific source models for self-training on unlabeled target data. SHOT performs information maximization and pseudo-label refinement~\cite{shot2020}, while AaD exploits neighborhood structure to improve target discriminability~\cite{22aad}. Subsequent methods improve adaptation robustness through domain-invariant parameter mining~\cite{2022dipe}, target prediction distribution search~\cite{24tpds}, advanced data augmentation~\cite{24sfda2}, and reliable pseudo-label selection~\cite{25tigm}. More recently, RGV-SFDA analyzes SFDA from a target-risk perspective~\cite{rsfda25}, whereas UCon-SFDA focuses on uncertainty control during adaptation~\cite{uconsfda25}. However, unlike unimodal SFDA that adapts task-specific visual source models, SFF-DA directly adapts pre-trained VLMs without source data or task-specific source models.

\myparagraph{Vision-Language Model Assisted SFDA.} Recent SFDA methods introduce CLIP as an external multimodal teacher to guide source-pretrained visual models, mainly through two paradigms: \textit{(1) Prediction-Level Calibration:} Co-learn++~\cite{25colearn} and DPC~\cite{24dpc} ensemble zero-shot priors within dual-branch frameworks, while ProDe~\cite{25prode} utilizes CLIP as a noisy agent for probabilistic denoising. \textit{(2) Structural-Level Alignment:} DIFO~\cite{difo2024} maximizes mutual information for alignment, while MMGA~\cite{25mmga} and DTKI~\cite{26dtki} enforce topological consistency to anchor features to semantic prototypes. However, these methods still rely on task-specific source models and therefore do not satisfy the SFF-DA setting.

\myparagraph{Multimodal Source-Fully-Free Domain Adaptation (SFF-DA).} 
Existing works mainly follow three paradigms: \textit{(1) Prompt Tuning}, which optimizes learnable prompts with frozen backbones (e.g., UPL~\cite{22upl}, TFUP-T~\cite{24tfupt}); \textit{(2) Visual Tuning}, which fine-tunes only the visual encoder to align with fixed text class embeddings (e.g., LaFTer~\cite{23lafter}, DPA~\cite{25dpa}); and \textit{(3) Joint Fine-tuning}, which optimizes both visual and text encoders for cross-modal alignment (e.g., ReCLIP~\cite{24reclip}, POUF~\cite{23pouf}, ImCapDA~\cite{25imcapda}). While prompt-based adaptation provides limited adaptability, visual tuning with fixed class embeddings can induce static semantic drift, whereas caption-guided joint fine-tuning may introduce dynamic semantic drift through unconstrained open-vocabulary semantics.

\section{DSSG Framework}

We propose DSSG to address the \textit{Dual Semantic Drift} dilemma in SFF-DA. It integrates VLM-generated captions, the DSG module, and Dynamic CMKD, while PAC extends the framework to DSSG-PAC by replacing iteration-wise class-prompt encoding with periodic prototype calibration.

\subsection{Preliminaries and Problem Definition}

\myparagraph{SFF-DA Setting:} Given an unlabeled target dataset $\mathcal{D}_t = \{x_i\}_{i=1}^{N_t}$, SFF-DA adapts a pre-trained vision-language model $\Theta_{clip}$ via unsupervised fine-tuning, without access to either source data $\mathcal{D}_s$ or source model $\Theta_{s}$.

\myparagraph{Revisiting CLIP:} CLIP comprises a visual encoder $E_V(\cdot;\theta_V)$ and a text encoder $E_T(\cdot;\theta_T)$ that are jointly pre-trained via large-scale contrastive learning to align visual and textual representations in a shared feature space. During inference, given a category set $\mathcal{C} = \{c_k\}_{k=1}^K$, fixed class embeddings $w_k = E_T(t_k)$ are computed from prompt templates $t_k$ (e.g., ``a photo of a $c_k$''). For an image $x_i$ with feature $f_i = E_V(x_i)$, the prediction probability is defined as:
\begin{equation}
	p(y_i=k|x_i) = \frac{\exp(\text{sim}(f_i, w_k) / \tau)}{\sum_{j=1}^{K} \exp(\text{sim}(f_i, w_j) / \tau)},
\end{equation}
where $\text{sim}(\cdot, \cdot)$ is the cosine similarity and $\tau$ is the temperature.

\subsection{Data Priors: Generative Instance-level Captions} 
In SFF-DA, directly fine-tuning CLIP induces evolutionary asymmetry between visual features and fixed textual class embeddings $\boldsymbol{w}_k$, resulting in static semantic drift. Although re-encoding class prompts at each iteration allows gradients to update the text encoder, such category-level guidance alone lacks instance-level target semantics and can suffer from catastrophic forgetting. Inspired by ImCapDA~\cite{25imcapda}, we leverage VLM-generated captions to provide target-specific guidance for joint vision--text adaptation. We use the pre-generated captions released by ImCapDA, with each target image $x_i\in\mathcal{D}_t$ paired with an instance-level caption $T_{\mathrm{cap},i}$. The caption generation process is formulated as: 
\begin{equation}
	T_{\mathrm{cap},i}=\text{VLM}(x_i),
\end{equation}
where the captions used in our experiments were generated by a frozen BLIP-3 model~\cite{xue2024xgen} with the prompt ``Describe what this image is in one sentence.'' Other VLMs can also be used as the caption generator. The generated captions $T_{\mathrm{cap},i}$ provide instance-level target-domain semantics, including object attributes and contextual information. Through image--caption contrastive alignment, gradients are propagated to $E_T(\cdot;\theta_T)$, enabling the text encoder to acquire domain-specific knowledge alongside visual adaptation. This target-aware textual guidance helps alleviate catastrophic forgetting, a key challenge in SFF-DA without source guidance.

\begin{figure*}[t]    
	\centering
	\begin{adjustbox}{width=0.99\linewidth,center}
		\includegraphics{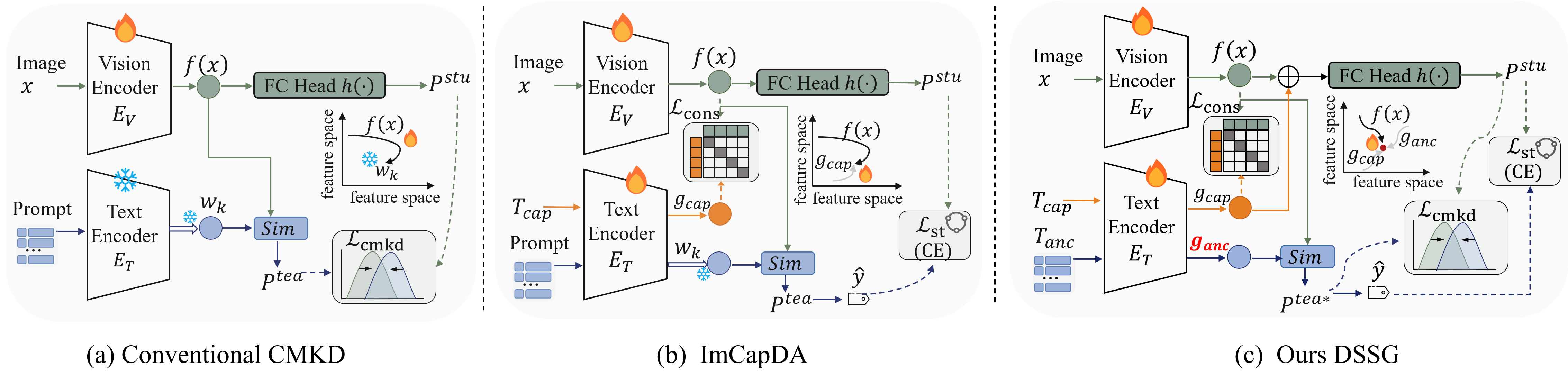}
	\end{adjustbox} 
	\caption{Illustration of (a) Conventional CMKD, (b) ImCapDA, and (c) Our DSSG framework.}
	\label{Fig-dssg-framework}
\end{figure*}

\subsection{Dual-Stream Guidance}
While $T_{\mathrm{cap}}$ introduces target-domain context, its unconstrained semantics can induce dynamic semantic drift, which, together with the aforementioned static semantic drift, constitutes the \textit{Dual Semantic Drift} dilemma. Unlike prior paradigms based on either fixed prompt templates~\cite{2024cmkd} or caption-only guidance~\cite{25imcapda}, DSG integrates a caption stream and a class-anchor stream in parallel. Specifically, for a target image $x_i$, its visual representation $f_i=E_V(x_i;\theta_V)$ is aligned with two complementary textual representations encoded by the shared text encoder $E_T(\cdot;\theta_T)$: 
\begin{equation}
	g_{anc} = E_T(T_{anc}; \theta_T), \quad g_{cap, i} = E_T(T_{cap, i}; \theta_T).
\end{equation} 
Here, $T_{\mathrm{anc}}=\{t_k\}_{k=1}^{K}$ denotes the class prompt set, which is re-encoded at each training iteration by the current text encoder to obtain $g_{\mathrm{anc}}$. The resulting $g_{\mathrm{anc}}$ preserves category semantics to mitigate dynamic semantic drift, whereas $g_{\mathrm{cap},i}$ captures instance-specific target-domain semantics from VLM-generated descriptions to alleviate static semantic drift. Together, the two streams balance global semantic stability and local adaptation while maintaining cross-modal alignment.

To jointly fine-tune ${\theta_V, \theta_T}$, we adopt a self-distillation paradigm following prior works~\cite{2024cmkd,25imcapda}, which employs a similarity-based teacher predictor and a linear student classifier $h(\cdot)$. Within this framework, we formulate objectives that exploit the adaptability of the caption stream while enforcing the semantic rigidity of the anchor stream:

\myparagraph{Caption-guided Contrastive Alignment.} 
To promote fine-grained cross-modal alignment, we employ a symmetric InfoNCE loss that maximizes the similarity of matched image--caption pairs $(f_i,g_{\mathrm{cap},i})$ while minimizing similarities to mismatched pairs. The objective comprises an image-to-text loss $\mathcal{L}_{i2t}$ and a text-to-image loss $\mathcal{L}_{t2i}$:
\begin{align}
	\mathcal{L}_{i2t} &= -\frac{1}{B}\sum_{i=1}^{B} \log \frac{\exp(\text{sim}(f_i,g_{\mathrm{cap},i})/\tau)}{\sum_{j=1}^{B}\exp(\text{sim}(f_i,g_{\mathrm{cap},j})/\tau)}, \\
	\mathcal{L}_{t2i} &= -\frac{1}{B}\sum_{i=1}^{B} \log \frac{\exp(\text{sim}(g_{\mathrm{cap},i},f_i)/\tau)}{\sum_{j=1}^{B}\exp(\text{sim}(g_{\mathrm{cap},i},f_j)/\tau)}.
\end{align}
where $B$ denotes the mini-batch size. The resulting contrastive objective is $\mathcal{L}_{\mathrm{con}}=0.5(\mathcal{L}_{i2t}+\mathcal{L}_{t2i})$.

\myparagraph{Dynamic Anchor-based Self-training.} 
To address the \textit{Dual Semantic Drift} described above, we update class anchors $\{g_{\mathrm{anc}}^k\}_{k=1}^{K}$ under the current $\theta_T$ to construct the dynamic teacher distribution:
\begin{equation}
	\begin{aligned}
		P^{tea} &= \text{Softmax}(\text{sim}(f_i, w_k) / \tau) &&\text{(Static)} \\
		P^{tea*} &= \text{Softmax}(\text{sim}(f_i, g_{anc}) / \tau) &&\text{(Dynamic)} ,
	\end{aligned}
\end{equation} 
where $P^{tea}$ provides dynamic guidance compared with the static baseline $P^{tea}$. Consequently, we derive pseudo-labels from the dynamic teacher: $\hat{y}_i=\arg\max_k P^{tea}(y=k|x_i)$. This dynamic anchor update is performed during training and introduces no additional cost at the inference stage.

Following ImCapDA~\cite{25imcapda}, we obtain the student prediction $P^{stu}$ by feeding the concatenated visual and caption features into a linear classifier $h(\cdot)$
\begin{equation}
	P^{stu}(y|x_i) = \text{Softmax}\big( h([f_i; g_{cap, i}]) \big).
\end{equation}
Based on the dynamic teacher prediction $P^{tea*}$, we formulate a confidence-thresholded self-training loss to transfer category-level anchor guidance to the student classifier. For each target sample, the teacher-derived pseudo-label $\hat{y}_i$ is used as supervision only when its prediction confidence exceeds $\gamma$:
\begin{equation}
	\mathcal{L}_{st}=-\frac{1}{B}\sum_{i=1}^{B}\mathbb{I}\big(P^{tea*}(\hat{y}_i|x_i)\ge\gamma\big)\log P^{stu}(\hat{y}_i|x_i),
\end{equation}
where $\mathbb{I}(\cdot)$ denotes the indicator function.

\subsection{Dynamic Cross-Modal Knowledge Distillation}  
To harness the  generalization capability of vision-language pre-training models, CMKD~\cite{2024cmkd} leverages the static semantic guidance from frozen text encoders to transfer generic knowledge to the unlabeled target domain. However, its static teacher guidance cannot adapt to the evolving representation space during training, limiting teacher--student consistency. To bridge this gap, Dynamic CMKD leverages the dynamic teacher $P^{tea*}$ and student $P^{stu}$ distributions obtained from the DSG module to enforce teacher--student consistency through a dynamic distillation objective.

\myparagraph{Optimization Objective.}
To quantify the prediction agreement between the student and the dynamic teacher, we define a consistency coefficient $coe=\mathrm{sg}(\exp(-\mathrm{KL}(P^{stu}\parallel P^{tea})))$, where $\mathrm{sg}(\cdot)$ denotes the stop-gradient operation.  o mitigate the risk of overconfidence associated with standard entropy minimization, we formulate the Dynamic CMKD objective using Gini impurity: 
\begin{equation}
	\begin{aligned}
		\mathcal{L}_{cmkd}^{*} &= \alpha \left[ \underbrace{coe \cdot \mathcal{G}(P^{stu})}_{\mathcal{L}_{task}} + \underbrace{(1 - coe) \cdot \mathcal{G}(P^{mix})}_{\mathcal{L}_{distill}} \right] \\
		&\quad + \underbrace{\beta \cdot \mathcal{G}(P^{tea*})}_{\mathcal{L}_{reg}} ,
	\end{aligned}
\end{equation}
where $\mathcal{G}(P)=1-|P|_2^2$ denotes the Gini impurity, and $P^{mix}=0.5(P^{stu}+P^{tea})$ denotes the averaged student–teacher distribution. Specifically, when $coe\to1$, $\mathcal{L}{task}$ encourages confident student predictions. When $coe\to0$, $\mathcal{L}{distill}$ increasingly relies on $P^{mix}$, allowing the dynamic teacher prior to regularize the student and reduce prediction bias. Meanwhile, $\mathcal{L}{reg}$ promotes a sharper teacher distribution to prevent over-smoothing.  We follow the original CMKD hyperparameter settings, with $\alpha=0.25$ and $\beta=0.025$.

\subsection{Overall Optimization Objective}   
For end-to-end optimization, DSSG integrates the caption stream $T_{\mathrm{cap}}$ and class-anchor stream $T_{\mathrm{anc}}$ to jointly fine-tune the visual encoder $\theta_V$ and text encoder $\theta_T$ via the following total objective:
\begin{equation}
	\mathcal{L}_{total} = \lambda_1 \mathcal{L}_{con} + \lambda_2 \mathcal{L}_{st} + \lambda_3 \mathcal{L}_{cmkd}^*, 
	\label{eq:overall_loss}
\end{equation}
where $\mathcal{L}_{\mathrm{con}}$ promotes caption-guided cross-modal alignment, $\mathcal{L}_{\mathrm{st}}$ performs confidence-thresholded self-training under dynamic teacher guidance, and $\mathcal{L}_{\mathrm{cmkd}}^*$ encourages prediction consistency between the anchor-based dynamic teacher and the visual–caption student classifier. Together, these objectives mitigate the \textit{Dual Semantic Drift}, while $\lambda_1$, $\lambda_2$, and $\lambda_3$ are trade-off hyperparameters.

\myparagraph{Total Gradient.} The visual and text encoders are jointly optimized through the aggregated gradients of the three objectives: 
$ \nabla_{\theta_V, \theta_T} \mathcal{L}_{total} = \lambda_1 \nabla_{\theta_V, \theta_T} \mathcal{L}_{con} + \lambda_2 \nabla_{\theta_V, \theta_T} \mathcal{L}_{st} + \lambda_3 \nabla_{\theta_V, \theta_T} \mathcal{L}_{cmkd}^*. $

%Crucially, the class anchor stream constrains the generated captions to prevent semantic divergence, effectively reconciling the capture of fine-grained domain knowledge with the preservation of global categorical consistency.

\subsection{From DSSG to DSSG-PAC: Prototype Anchor Calibration} 
In DSSG, the class-anchor stream re-encodes all class prompts at each training iteration to obtain class text features from the current text encoder. Similarly, prompt-learning-based DA methods~\cite{dapl2023,ge2022adclip} construct class prompts with learnable context token embeddings, which must still pass through the text encoder to produce the final class features. Their text-side encoding pipelines can be summarized as:
\begin{equation}
	\begin{aligned}
		&\text{DSSG:}
		\\[-2pt]
		&\quad
		T_{\mathrm{anc}}
		\rightarrow
		\text{tokenized\_prompts}
		\in
		\mathbb{N}^{K\times 77}
		\\
		&\quad
		{}\rightarrow
		\text{token\_embedding}
		\in
		\mathbb{R}^{K\times 77\times d_{\mathrm{tok}}}
		\\
		&\quad
		{}\rightarrow
		E_T
		\rightarrow
		g_{\mathrm{anc}}
		\in
		\mathbb{R}^{K\times d},
		\\[4pt]
		&\text{Prompt Learning:}
		\\[-2pt]
		&\quad
		\left\{
		[v_1,\ldots,v_M,c_k]
		\right\}_{k=1}^{K}
		\in
		\mathbb{R}^{K\times(M+1)\times d_{\mathrm{tok}}}
		\\
		&\quad
		\xrightarrow{\mathrm{padding}_{77}}
		\text{token\_embedding}
		\in
		\mathbb{R}^{K\times 77\times d_{\mathrm{tok}}}
		\\
		&\quad
		{}\rightarrow
		E_T
		\rightarrow
		g_{\mathrm{pl}}
		\in
		\mathbb{R}^{K\times d},
	\end{aligned}
	\label{eq:text_encoding_pipelines}
\end{equation}
where $[v_1,\ldots,v_M]$ denotes the shared learnable context embeddings, $c_k$ denotes the fixed embedding corresponding to the $k$-th class name, and $d_{\mathrm{tok}}$ and $d$ denote the token-embedding and text-feature dimensions, respectively. Although DSSG uses fixed class prompts whereas prompt-learning methods employ learnable contexts, both encode $K$ class prompts through $E_T$ at every iteration, incurring an iteration-level text-encoding cost that scales with the number of classes. We further instantiate prompt learning in DSSG as DSSG-PL, which replaces the fixed class prompts with learnable context embeddings.

\myparagraph{Prototype Anchor Calibration.} 
To avoid redundant encoding at every iteration, instead of optimizing class anchors through prompt learning in the token-embedding space, we directly model the text-encoded class features as periodically calibrated prototype anchors in the text feature space, giving rise to DSSG with Prototype Anchor Calibration (DSSG-PAC). Let $\Delta_{\mathrm{cal}}$ denote the calibration interval measured in epochs, and $t_m$ the training iteration at which the $m$-th calibration is performed. At $t_m$, the text encoder generates the class prototypes using its current parameters as
\begin{equation}
g_{\mathrm{pac}}^{(m)}
=
\operatorname{sg}\left[
E_T\left(
T_{\mathrm{anc}};
\theta_T^{(t_m)}
\right)
\right]
\in
\mathbb{R}^{K\times d},
\end{equation}
where $\theta_T^{(t_m)}$ denotes the current text-encoder parameters and $\operatorname{sg}(\cdot)$ stops gradient propagation. These prototypes are fixed and reused as class anchors for $t_m\leq t<t_{m+1}$ without propagating gradients to $E_T$, whereas $E_T$ remains optimized through the $T_{\mathrm{cap}}$ branch and its updated parameters are used to regenerate the prototypes at the next calibration. We set $\Delta_{\mathrm{cal}}=1$ by default, corresponding to one calibration per epoch, and evaluate other intervals in the ablation study.  

Specifically, PAC replaces the $g_{\mathrm{anc}}$ generated at every iteration in the class-anchor stream with the periodically calibrated and reused $g_{\mathrm{pac}}^{(m)}$, reducing text-encoding overhead while preserving the adaptability of the class prototypes to changes in the text feature space. Given the visual feature $f_i$ of a target image $x_i$, the corresponding teacher distribution is
\begin{equation}
p_{\mathrm{tea}}^{*}(y=k\mid x_i)
=
\frac{
	\exp\left(
	\operatorname{cos}\left(
	f_i,
	g_{\mathrm{pac},k}^{(m)}
	\right)/\tau
	\right)
}{
	\sum_{j=1}^{K}
	\exp\left(
	\operatorname{cos}\left(
	f_i,
	g_{\mathrm{pac},j}^{(m)}
	\right)/\tau
	\right)
}.
\end{equation}
All other components and optimization objectives of DSSG remain unchanged. 

For $N_{\mathrm{ep}}$ training epochs with $I$ iterations per epoch, DSSG performs $N_{\mathrm{ep}}I$ class-prompt encoding operations, each involving all $K$ classes, resulting in a text-side encoding workload that scales with $N_{\mathrm{ep}}IK$. Thus, the text-side overhead increases with both the target-domain size and the number of classes. In contrast, DSSG-PAC performs only approximately $\lceil N_{\mathrm{ep}}/\Delta_{\mathrm{cal}}\rceil$ such operations and maintains a $K\times d$ prototype matrix between calibrations.

\myparagraph{Risk Analysis for Source-Fully-Free Adaptation}
\label{sec:sffda_theory}

Unlike conventional UDA bounds that depend on source samples or a task-specific source hypothesis, neither is available in SFF-DA. We therefore characterize the target risk relative to the semantic teacher that is actually available during adaptation. Let $q(x)=P_{\mathrm{tea}}^{*}(\cdot\mid x)$ and $p(x)=P_{\mathrm{stu}}(\cdot\mid x)$, and let $h_q$ and $h_p$ be their induced top-1 classifiers. For the unknown target distribution $Q$, define the teacher risk $\eta_Q=\Pr_{(x,y)\sim Q}[h_q(x)\neq y]$ and the student risk $R_Q(h_p)=\Pr_{(x,y)\sim Q}[h_p(x)\neq y]$. Let $m_q(x)=q_{(1)}(x)-q_{(2)}(x)$ denote the teacher's top-1 probability margin and define
\begin{equation}
	\Psi(p,q;x)=
	\begin{cases}
		\displaystyle \min\!\left\{1,
		\frac{2\lVert p(x)-q(x)\rVert_2^2}{m_q(x)^2}\right\},
		&m_q(x)>0,\\
		1,&m_q(x)=0,
	\end{cases}
	\label{eq:certificate}
\end{equation}
with $\mathcal C_Q(p,q)=\mathbb E_{x\sim Q_X}\Psi(p,q;x)$.

\noindent\textbf{Theorem 1 (Certified SFF-DA risk bound).}
For any semantic-teacher and student distributions $q$ and $p$,
\begin{equation}
	\Pr_{Q_X}[h_p(x)\neq h_q(x)]\leq \mathcal C_Q(p,q),
	\label{eq:disagreement_certificate}
\end{equation}
and
\begin{equation}
	\max\{0,\eta_Q-\mathcal C_Q(p,q)\}
	\leq R_Q(h_p)\leq
	\min\{1,\eta_Q+\mathcal C_Q(p,q)\}.
	\label{eq:sffda_risk_bound}
\end{equation}
The proof is provided in Appendix~\ref{app:risk_proof}.

The bound separates the irreducible semantic quality of the dynamic teacher, $\eta_Q$, from a label-free, margin-normalized consistency certificate, $\mathcal C_Q$. In DSSG, the class-anchor stream is designed to reduce semantic drift in $q$, while Dynamic CMKD contains an explicitly disagreement-adaptive term proportional to $\lVert p-q\rVert_2^2$, which is aligned with the certificate in Eq.~\eqref{eq:certificate}. We stress that a large teacher margin measures low predictive ambiguity rather than correctness; consequently, reduction of $\eta_Q$ is an empirical property assessed through oracle pseudo-label analysis, not an assumption-free consequence of the bound.

\begin{comment}
	\noindent\textbf{Corollary 1 (DSSG--PAC performance certificate).}
	Let $h_D$ and $h_P$ denote the final DSSG and DSSG-PAC classifiers. Their target-risk gap obeys
	\begin{equation}
		\big|R_Q(h_P)-R_Q(h_D)\big|
		\leq \Pr_{x\sim Q_X}[h_P(x)\neq h_D(x)].
		\label{eq:pac_dssg_bound}
	\end{equation}
	Thus, final-model agreement gives a direct label-free certificate on the largest possible accuracy gap between PAC and DSSG. This relative guarantee is distinct from the within-model certificate in Eq.~\eqref{eq:sffda_risk_bound} and does not assert unconditional superiority on an arbitrary target labeling function.
\end{comment}
\section{Experiments}

\subsection{Experiments Setting}

\myparagraph{Datasets.} Experiments are evaluated on four benchmarks:  
(i) \textit{Office-31}, 4,110 images in 31 classes across three domains: Amazon (A), Webcam (W), and DSLR (D);
(ii) \textit{Office-Home}, 15,588 images in 65 classes across four domains: Art (A), Clipart (C), Product (P), and Real-World (R);
(iii) \textit{Mini-DomainNet}, a DomainNet subset with 140,006 images in 126 classes across four domains: Clipart (C), Painting (P), Real (R), and Sketch (S);
(iv) \textit{VisDA}, a synthetic-to-real benchmark with 55,388 target images across 12 classes.

\myparagraph{Baselines.} We evaluate our method against representative baselines from three categories: 
(i) \textit{Unimodal Source-Free DA} (U), including SHOT~\cite{shot2020}, AaD~\cite{22aad}, DIPE~\cite{2022dipe}, DSiT~\cite{23dsit}, SF(DA)$^2$~\cite{24sfda2}, and TPDS~\cite{24tpds}; 
(ii) \textit{Multimodal Source-Free DA} (U+M), employing an external multimodal teacher to guide the source-supervised unimodal student, represented by DIFO-C~\cite{24difo}, ProDe~\cite{25prode}, MMGA~\cite{25mmga}, Co-Learn~\cite{25colearn}, DTKI~\cite{26dtki}, and DPC~\cite{24dpc}; and 
(iii) \textit{Multimodal Source-Fully-Free DA} (M), such as UPL~\cite{22upl}, POUF~\cite{23pouf}, TFUP-T~\cite{24tfupt}, ReCLIP~\cite{24reclip}, and ImCapSFDA~\cite{25imcapda}.

\myparagraph{Implementation Details.} In the strict SFF-DA setting, target-domain labels are unavailable for model selection; therefore, we report Last Epoch accuracy as the primary evaluation metric. All experiments are conducted using three random seeds $\{2020,2026,2027\}$, and the averaged results are reported. Detailed experimental configurations are provided in Appendix~\ref{app:exp_config}.

\begin{table*}[t] 
	\caption{Classification accuracy (\%) on Office-Home. $\dagger$ denotes a multimodal teacher with its backbone, ``--'' means not reported.}
	
	\label{Tab-sfda-oh}
	\centering
	\begin{adjustbox}{width=1.0\linewidth,center}
		\begin{tabular}{l@{\hskip 1pt}l@{\hskip 3pt}l@{\hskip 4pt}c|c|c|c|c|c|c|c|c|c|c|c|c|c|c}
			\hline
			\rule{0pt}{10pt}\multirow{2}{*}{Methods} & \multirow{2}{*}{Venues} & \multirow{2}{*}{Prot.} & \multirow{2}{*}{Mark} & C & P & R & A & P & R & A & C & R & A & C & P & \multirow{2}{*}{\makecell{Avg.\\Last}} & \multirow{2}{*}{\makecell{Avg.\\Best}} \\ \cline{5-16}
			\rule{0pt}{10pt} & & & & \mc{A} & \mc{C} & \mc{P} & \mc{R} & & \\ \hline
			\multicolumn{1}{c}{\textit{ResNet-50:}} & \\
			SHOT~\cite{shot2020} & \citeye{ICML20} & SF & U & 68.0 & 67.4 & 73.3 & 57.1 & 54.9 & 58.8 & 78.1 & 78.2 & 84.3 & 81.5 & 78.1 & 82.2 & 71.8 & -- \\
			SF(DA)$^2$~\cite{24sfda2} & \citeye{ICLR24} & SF & U & 69.5 & 66.5 & 73.3 & 57.8 & 57.2 & 60.2 & 80.2 & 79.2 & 83.8 & 81.5 & 79.4 & 82.1 & 72.6 & -- \\
			TIGM~\cite{25tigm} & \citeye{CVPR25} & SF & U & 69.3 & 68.1 & 74.6 & 61.2 & 58.8 & 62.4 & 80.9 & 81.2 & 85.7 & 82.7 & 81.4 & 83.4 & 74.1 & -- \\
			\hdashline
			\rule{0pt}{10pt}DIFO-V$^\dagger$$_{B/32}$~\cite{24difo} & \citeye{CVPR24} & SF & U+M & 82.5 & 80.9 & 83.4 & 70.6 & 70.1 & 70.5 & 90.6 & 90.6 & 91.2 & 88.8 & 88.8 & 88.9 & 83.1 & -- \\
			ProDe-V$^\dagger$$_{B/32}$~\cite{25prode} & \citeye{ICLR25} & SF & U+M & 82.5 & 82.5 & 83.0 & 72.7 & 72.5 & 72.6 & 92.3 & 91.5 & 92.2 & 90.5 & 90.7 & 90.8 & 84.5 & -- \\
			MMGA$^\dagger$$_{R/50}$~\cite{25mmga} & \citeye{PR25} & SF & U+M & 77.2 & 75.9 & 79.4 & 65.8 & 65.3 & 66.1 & 84.5 & 86.9 & 88.7 & 85.9 & 86.1 & 86.1 & 79.0 & -- \\
			Co-learn$^\dagger$$_{L/14}$~\cite{25colearn} & \citeye{IJCV25} & SF & U+M & 77.1 & 76.6 & 82.0 & 77.2 & 72.5 & 79.6 & 90.4 & 88.1 & 93.0 & 91.0 & 90.0 & 90.1 & 84.0 & -- \\
			DTKI$^\dagger$$_{B/32}$~\cite{26dtki} & \citeye{IPM26} & SF & U+M & 82.3 & 82.8 & 82.7 & 72.0 & 71.1 & 71.7 & 91.7 & 91.9 & 91.6 & 90.0 & 90.0 & 90.3 & 84.0 & -- \\
			\cline{5-18}
			\rule{0pt}{10pt}CLIP-RN50~\cite{radford2021learning} & \citeye{ICML21} & SFF & M & \mc{71.5} & \mc{54.5} & \mc{80.0} & \mc{82.3} & 72.1 & -- \\
			ImCapSFDA~\cite{25imcapda} & \citeye{ESWA26} & SFF & M & \mc{81.6 $\pm$ 0.11} & \mc{76.7 $\pm$ 0.08} & \mc{88.6 $\pm$ 0.08} & \mc{89.8 $\pm$ 0.14} & 84.2 $\pm$ 0.04 & 84.3 $\pm$ 0.05 \\
			\rowcolor{gray!25}DSSG & \citeye{Ours} & SFF & M & \mc{\textbf{84.5} $\pm$ 0.20} & \mc{\textbf{83.7} $\pm$ 0.31} & \mc{\textbf{94.5} $\pm$ 0.20} & \mc{\textbf{92.4} $\pm$ 0.11} & \textbf{88.8} $\pm$ 0.06 & \textbf{89.2} $\pm$ 0.05 \\
			\rowcolor{gray!25}DSSG-PAC & \citeye{Ours} & SFF & M & \mc{\textbf{84.5} $\pm$ 0.35} & \mc{83.5 $\pm$ 0.27} & \mc{94.2 $\pm$ 0.11} & \mc{\textbf{92.4} $\pm$ 0.15} & 88.7 $\pm$ 0.09 & 89.1 $\pm$ 0.11 \\ 
			\hline
			\hline
			\multicolumn{1}{c}{\textit{ViT-B/16:}} & \\
			SHOT~\cite{shot2020} & \citeye{ICML20} & SF & U & 76.6 & 76.3 & 80.4 & 67.1 & 65.3 & 66.7 & 83.5 & 83.4 & 83.4 & 85.5 & 83.7 & 85.3 & 78.1 & -- \\
			DIPE~\cite{2022dipe} & \citeye{CVPR22} & SF & U & 77.1 & 75.3 & 81.6 & 66.0 & 63.3 & 67.7 & 80.8 & 83.5 & 89.6 & 85.6 & 83.4 & 85.1 & 78.2 & -- \\
			DSiT~\cite{23dsit} & \citeye{ICCV23} & SF & U & 80.7 & 77.9 & 82.4 & 69.2 & 67.9 & 68.3 & 83.5 & 86.1 & 89.8 & 87.3 & 86.2 & 86.6 & 80.5 & -- \\
			DPC$^\dagger$$_{B/16}$~\cite{24dpc} & \citeye{IJCAI24} & SF & U+M & 86.3 & 86.6 & 87.6 & 71.1 & 74.1 & 75.0 & 87.1 & 90.9 & 91.8 & 91.3 & 91.6 & 91.8 & 85.4 & -- \\
			\cline{5-18}
			\rule{0pt}{10pt}CLIP-ViT/B-16~\cite{radford2021learning} & \citeye{ICML21} & SFF & M & \mc{82.3} & \mc{71.3} & \mc{89.2} & \mc{89.9} & 83.2 & -- \\
			UPL~\cite{22upl} & \citeye{ArXiv22} & SFF & M & \mc{83.3} & \mc{67.7} & \mc{91.5} & \mc{90.7} & 83.3 & -- \\
			POUF~\cite{23pouf} & \citeye{ICML23} & SFF & M & \mc{86.2} & \mc{73.8} & \mc{92.7} & \mc{91.7} & 86.1 & -- \\
			TFUP-T~\cite{24tfupt} & \citeye{ArXiv24} & SFF & M & \mc{86.0} & \mc{74.2} & \mc{93.1} & \mc{91.7} & 86.3 & -- \\
			ReCLIP~\cite{24reclip} & \citeye{WACV24} & SFF & M & \mc{86.1} & \mc{76.0} & \mc{93.9} & \mc{92.0} & 87.0 & -- \\
			ImCapSFDA~\cite{25imcapda} & \citeye{ESWA26} & SFF & M & \mc{87.5 $\pm$ 0.05} & \mc{87.0 $\pm$ 0.05} & \mc{94.4 $\pm$ 0.01} & \mc{93.9 $\pm$ 0.06} & 90.7 $\pm$ 0.03 & 90.8 $\pm$ 0.01 \\
			\rowcolor{gray!25}DSSG & \citeye{Ours} & SFF & M & \mc{90.0 $\pm$ 0.04} & \mc{\textbf{89.2} $\pm$ 0.15} & \mc{\textbf{97.3} $\pm$ 0.05} & \mc{\textbf{94.6} $\pm$ 0.09} & \textbf{92.8} $\pm$ 0.02 & \textbf{92.8} $\pm$ 0.01 \\
			\rowcolor{gray!25}DSSG-PAC & \citeye{Ours} & SFF & M & \mc{\textbf{90.1} $\pm$ 0.08} & \mc{\textbf{89.2} $\pm$ 0.12} & \mc{\textbf{97.3} $\pm$ 0.10} & \mc{\textbf{94.6} $\pm$ 0.01} & \textbf{92.8} $\pm$ 0.02 & \textbf{92.8} $\pm$ 0.01 \\ 
			\hline
		\end{tabular}
	\end{adjustbox}
\end{table*}

\begin{table}[t]
	\centering
	\caption{Classification accuracy (\%) on Office-31. $\dagger$ denotes a multimodal teacher with its backbone, ``--'' means not reported.}
	\label{Tab-sfda-o31}
	
	\begin{adjustbox}{width=1\linewidth,center} 
		\begin{tabular}{l@{\hskip 1pt}l@{\hskip 4pt}c|cc|cc|cc|c|c}
			\hline
			\rule{0pt}{10pt}\multirow{2}{*}{Methods} & \multirow{2}{*}{Prot.} & \multirow{2}{*}{Mark} & D & W & A & W & A & D & \multirow{2}{*}{\makecell{Avg.\\Last}} & \multirow{2}{*}{\makecell{Avg.\\Best}} \\ \cline{4-9}
			\rule{0pt}{10pt} & & & \multicolumn{2}{c|}{A} & \multicolumn{2}{c|}{D} & \multicolumn{2}{c|}{W} & & \\ \hline
			\multicolumn{1}{c}{\textit{ResNet-50:}} & \\
			SHOT~\cite{shot2020} & SF & U & 74.7 & 77.8 & 94.0 & 99.8 & 90.1 & 98.4 & 88.6 & -- \\
			SF(DA)$^2$~\cite{24sfda2} & SF & U & 75.7 & 76.8 & 95.8 & 99.8 & 92.1 & 99.0 & 89.9 & -- \\
			\hdashline
			\rule{0pt}{10pt}DIFO-V$^\dagger$$_{B/32}$~\cite{24difo} & SF & U+M & 83.0 & 83.2 & 97.2 & 98.8 & 95.5 & 97.2 & 92.5 & -- \\
			ProDe-V$^\dagger$$_{B/32}$~\cite{25prode} & SF & U+M & 83.1 & 82.5 & 96.8 & 99.8 & 96.4 & 97.0 & 92.6 & -- \\
			%Co-learn$^\dagger$$_{L/14}$~\cite{25colearn} & SF & U+M & 85.3 & 83.2 & 99.2 & 100.0 & 99.7 & 99.1 & 94.4 & -- \\
			DTKI$^\dagger$$_{B/32}$~\cite{26dtki} & SF & U+M & 83.2 & 83.1 & 98.2 & 99.2 & 95.1 & 98.8 & 92.9 & -- \\
			\cline{4-11}
			\rule{0pt}{10pt}CLIP-RN50~\cite{radford2021learning} & SFF & M & \multicolumn{2}{c|}{73.2} & \multicolumn{2}{c|}{72.5} & \multicolumn{2}{c|}{70.1} & 71.9 & -- \\
			ImCapSFDA~\cite{25imcapda} & SFF & M & \multicolumn{2}{c|}{77.8$\pm$0.34} & \multicolumn{2}{c|}{88.0$\pm$0.40} & \multicolumn{2}{c|}{86.2$\pm$0.08} & 84.0$\pm$0.04 & 84.2$\pm$0.09 \\
			\rowcolor{gray!25}DSSG & SFF & M & \multicolumn{2}{c|}{\textbf{83.9}$\pm$0.41} & \multicolumn{2}{c|}{\textbf{93.0}$\pm$1.25} & \multicolumn{2}{c|}{\textbf{94.0}$\pm$0.38} & \textbf{90.3}$\pm$0.67 & \textbf{91.0}$\pm$0.17 \\
			\rowcolor{gray!25}DSSG-PAC & SFF & M & \multicolumn{2}{c|}{83.5$\pm$0.32} & \multicolumn{2}{c|}{92.0$\pm$1.23} & \multicolumn{2}{c|}{93.5$\pm$0.82} & 89.7$\pm$0.49 & 90.4$\pm$0.16 \\
			\hline
			\hline
			\multicolumn{1}{c}{\textit{ViT-B/16:}} & \\
			SHOT~\cite{shot2020} & SF & U & 79.4 & 80.2 & 95.3 & 100.0 & 94.3 & 99.0 & 91.4 & -- \\
			DIPE~\cite{2022dipe} & SF & U & 77.5 & 77.1 & 94.8 & 100.0 & 95.5 & 98.5 & 90.5 & -- \\
			DSiT~\cite{23dsit} & SF & U & 81.7 & 81.9 & 98.0 & 100.0 & 97.2 & 99.1 & 93.0 & -- \\
			DPC$^\dagger$$_{B/16}$~\cite{24dpc} & SF & U+M & 83.9 & 83.7 & 96.9 & 100.0 & 97.0 & 98.2 & 93.3 & -- \\
			\cline{4-11}
			\rule{0pt}{10pt}CLIP-ViT-B/16~\cite{radford2021learning} & SFF & M & \multicolumn{2}{c|}{78.9} & \multicolumn{2}{c|}{79.9} & \multicolumn{2}{c|}{76.9} & 78.6 & -- \\
			UPL~\cite{22upl} & SFF & M & \multicolumn{2}{c|}{81.4} & \multicolumn{2}{c|}{82.6} & \multicolumn{2}{c|}{83.6} & 82.5 & -- \\
			POUF~\cite{23pouf} & SFF & M & \multicolumn{2}{c|}{84.4} & \multicolumn{2}{c|}{91.1} & \multicolumn{2}{c|}{91.3} & 88.9 & -- \\
			TFUP-T~\cite{24tfupt} & SFF & M & \multicolumn{2}{c|}{84.8} & \multicolumn{2}{c|}{90.8} & \multicolumn{2}{c|}{93.2} & 89.6 & -- \\
			ImCapSFDA~\cite{25imcapda} & SFF & M & \multicolumn{2}{c|}{85.2$\pm$0.11} & \multicolumn{2}{c|}{91.8$\pm$0.31} & \multicolumn{2}{c|}{91.6$\pm$0.22} & 89.5$\pm$0.10 & 89.6$\pm$0.13 \\
			\rowcolor{gray!25}DSSG & SFF & M & \multicolumn{2}{c|}{\textbf{87.6}$\pm$0.13} & \multicolumn{2}{c|}{\textbf{98.4}$\pm$0.00} & \multicolumn{2}{c|}{\textbf{97.2}$\pm$0.00} & \textbf{94.4}$\pm$0.04 & \textbf{94.8}$\pm$0.06 \\
			\rowcolor{gray!25}DSSG-PAC & SFF & M & \multicolumn{2}{c|}{85.2$\pm$0.12} & \multicolumn{2}{c|}{97.5$\pm$0.23} & \multicolumn{2}{c|}{96.7$\pm$0.82} & 93.1$\pm$0.18 & 93.4$\pm$0.20 \\
			\hline
		\end{tabular}
	\end{adjustbox}
\end{table}

\begin{table*}[t]
	\centering 
	\caption{Classification accuracy (\%) on Mini-DomainNet. $\dagger$ denotes a multimodal teacher with its backbone, ``--'' means not reported.}
	\label{Tab-sfda-mini}
	\begin{adjustbox}{width=1.0\linewidth,center}
		\begin{tabular}{l@{\hskip 1pt}c@{\hskip 3pt}c@{\hskip 4pt}c|c|c|c|c|c|c|c|c|c|c|c|c|c|c}
			\hline
			\rule{0pt}{10pt}\multirow{2}{*}{Methods} & \multirow{2}{*}{Venues} & \multirow{2}{*}{Prot.} & \multirow{2}{*}{Mark} & P & R & S & C & R & S & C & P & S & C & P & R & \multirow{2}{*}{\makecell{Avg.\\Last}} & \multirow{2}{*}{\makecell{Avg.\\Best}}  \\ 
			
			\cline{5-16}
			\rule{0pt}{10pt} & & & & \multicolumn{3}{c|}{C} & \multicolumn{3}{c|}{P} & \multicolumn{3}{c|}{R} & \multicolumn{3}{c|}{S} & & \\
			\hline
			\multicolumn{1}{c}{\textit{ResNet-50:}} \\
			SHOT~\cite{shot2020} & \citeye{ICML20} & SF & U & 67.9 & 67.7 & 70.2 & 63.5 & 67.6 & 64.0 & 78.2 & 81.3 & 78.0 & 59.5 & 61.7 & 57.8 & 68.1 & -- \\
			TPDS~\cite{24tpds} & \citeye{IJCV24} & SF & U & 65.6 & 66.4 & 68.6 & 62.9 & 67.0 & 64.3 & 77.1 & 79.0 & 75.3 & 59.8 & 61.5 & 68.2 & 67.1 & -- \\
			\hdashline
			\rule{0pt}{10pt}DIFO-V$^\dagger$$_{B/32}$~\cite{24difo} & \citeye{CVPR24} & SF & U+M & 80.0 & 80.8 & 80.5 & 76.6 & 77.3 & 76.7 & 87.2 & 87.4 & 87.3 & 74.9 & 75.6 & 75.5 & 80.0 & -- \\
			ProDe-V$^\dagger$$_{B/32}$~\cite{25prode} & \citeye{ICLR25} & SF & U+M & 85.0 & 85.5 & 85.5 & 83.2 & 83.1 & 83.4 & 92.4 & 92.3 & 92.4 & 79.0 & 79.3 & 79.1 & 85.0 & -- \\
			MMGA$^\dagger$$_{R/50}$~\cite{25mmga} & \citeye{PR25} & SF & U+M & 74.6 & 75.9 & -- & -- & 76.1 & 75.6 & -- & 88.3 & -- & 69.9 & -- & 70.9 & 75.9 & -- \\
			Co-learn$^\dagger$$_{L/14}$~\cite{25colearn} & \citeye{IJCV25} & SF & U+M & 78.9 & 85.4 & 81.2 & 75.1 & 79.1 & 73.8 & 86.5 & 86.7 & 84.4 & 78.5 & 76.8 & 76.7 & 80.3 & -- \\
			DTKI$^\dagger$$_{B/32}$~\cite{26dtki} & \citeye{IPM26} & SF & U+M & 82.4 & 80.7 & 82.3 & 77.5 & 78.7 & 78.9 & 88.9 & 88.0 & 88.3 & 76.1 & 76.9 & 74.6 & 81.1 & -- \\
			\cline{5-18}
			\rule{0pt}{10pt}CLIP-RN~\cite{radford2021learning} & \citeye{ICML21} & SFF & M & \mc{68.4} & \mc{69.3} & \mc{86.2} & \mc{65.9} & 72.5 & -- \\
			ImCapSFDA~\cite{25imcapda} & \citeye{ESWA26} & SFF & M & \mc{84.2 $\pm$ 0.15} & \mc{84.0 $\pm$ 0.03} & \mc{91.9 $\pm$ 0.02} & \mc{83.8 $\pm$ 0.06} & 86.0 $\pm$ 0.05 & 86.0 $\pm$ 0.06 \\
			\rowcolor{gray!25}DSSG & \citeye{Ours} & SFF & M & \mc{\textbf{86.8} $\pm$ 0.00} & \mc{\textbf{84.5} $\pm$ 0.14} & \mc{\textbf{93.5} $\pm$ 0.06} & \mc{84.2 $\pm$ 0.21} & \textbf{87.3} $\pm$ 0.08 & \textbf{87.3} $\pm$ 0.07 \\
			\rowcolor{gray!25}DSSG-PAC & \citeye{Ours} & SFF & M & \mc{86.5 $\pm$ 0.23} & \mc{84.4 $\pm$ 0.12} & \mc{93.4 $\pm$ 0.05} & \mc{\textbf{84.3} $\pm$ 0.17} & 87.2 $\pm$ 0.02 & 87.2 $\pm$ 0.01 \\
			\hline
			\hline
			\multicolumn{1}{c}{\textit{ViT-B/16:}} & \\
			SHOT~\cite{shot2020} & \citeye{ICML20} & SF & U & 68.9 & 72.3 & 74.0 & 64.8 & 70.6 & 69.2 & 82.3 & 84.0 & 83.6 & 63.1 & 62.7 & 61.7 & 71.4 & -- \\
			AaD~\cite{22aad} & \citeye{NIPS22} & SF & U & 70.4 & 74.6 & 76.4 & 66.8 & 72.1 & 71.2 & 81.0 & 84.0 & 82.8 & 63.8 & 65.4 & 63.8 & 72.7 & -- \\
			DPC$^\dagger$$_{B/16}$~\cite{24dpc} & \citeye{IJCAI24} & SF & U+M & 89.7 & 86.1 & 87.2 & 82.5 & 82.9 & 84.8 & 89.6 & 91.2 & 89.2 & 82.1 & 80.9 & 81.4 & 85.6 & -- \\
			\cline{5-18}
			\rule{0pt}{10pt}CLIP-ViT/B-16~\cite{radford2021learning} & \citeye{ICML21} & SFF & M & \mc{84.2} & \mc{82.0} & \mc{91.5} & \mc{80.9} & 84.7 & -- \\
			ImCapSFDA~\cite{25imcapda} & \citeye{ESWA26} & SFF & M & \mc{90.5 $\pm$ 0.03} & \mc{89.4 $\pm$ 0.04} & \mc{94.2 $\pm$ 0.02} & \mc{\textbf{89.1} $\pm$ 0.06} & 90.8 $\pm$ 0.03 & 90.8 $\pm$ 0.03 \\
			\rowcolor{gray!25}DSSG & \citeye{Ours} & SFF & M & \mc{\textbf{90.8} $\pm$ 0.07} & \mc{\textbf{89.5} $\pm$ 0.07} & \mc{\textbf{95.1} $\pm$ 0.01} & \mc{89.0 $\pm$ 0.04} & \textbf{91.1} $\pm$ 0.02 & \textbf{91.1} $\pm$ 0.02 \\
			\rowcolor{gray!25}DSSG-PAC & \citeye{Ours} & SFF & M & \mc{\textbf{90.8} $\pm$ 0.10} & \mc{89.4 $\pm$ 0.03} & \mc{\textbf{95.1} $\pm$ 0.01} & \mc{89.0 $\pm$ 0.02} & \textbf{91.1} $\pm$ 0.03 & \textbf{91.1} $\pm$ 0.03 \\
			\hline
		\end{tabular}
	\end{adjustbox}
\end{table*}

\begin{table*}[t]
	\caption{Per-category classification accuracy (\%) on VisDA. $\dagger$ denotes a multimodal teacher with its backbone, ``--'' means not reported.}
	\label{Tab-sfda-visda-per}
	\centering
	\begin{adjustbox}{width=1.0\linewidth,center}
		\begin{tabular}{l@{\hskip 1pt}l@{\hskip 4pt}l@{\hskip 3pt}c|cccccccccccc|c|c}
			\hline
			\rule{0pt}{10pt}Methods & Venues & Prot. & Mark & Pln & Bcy & Bus & Car & Hor & Kni & Mcy & Per & Plt & Skt & Trn & Trk & Avg. Last & Avg. Best \\ \hline
			\multicolumn{1}{c}{\textit{ResNet-101:}} & \\
			SHOT~\cite{shot2020} & \citeye{ICML20} & SF & U & 94.3 & 88.5 & 80.1 & 57.3 & 93.1 & 94.9 & 80.7 & 80.3 & 91.5 & 89.1 & 86.3 & 58.2 & 82.9 & -- \\
			SF(DA)$^2$~\cite{24sfda2} & \citeye{ICLR24} & SF & U & 96.8 & 89.3 & 82.9 & 81.4 & 96.8 & 95.7 & 90.4 & 81.3 & 95.5 & 93.7 & 88.5 & 64.7 & 88.1 & -- \\
			\hdashline
			\rule{0pt}{10pt}DIFO-V$^\dagger$$_{B/32}$~\cite{24difo} & \citeye{CVPR24} & SF & U+M & 97.5 & 89.0 & 90.8 & \textbf{83.5} & 97.8 & 97.3 & 93.2 & 83.5 & 95.2 & 96.8 & 93.7 & 65.9 & 90.3 & -- \\
			ProDe-V$^\dagger$$_{B/32}$~\cite{25prode} & \citeye{ICLR25} & SF & U+M & 98.3 & 92.4 & 86.6 & 80.5 & 98.1 & 98.0 & 92.3 & 84.3 & 94.7 & 97.0 & 94.1 & \textbf{75.6} & \textbf{91.0} & -- \\
			MMGA$^\dagger$$_{R/50}$~\cite{25mmga} & \citeye{PR25} & SF & U+M & 97.2 & 85.0 & 90.2 & 82.9 & 96.8 & 97.1 & 93.4 & 84.6 & 90.7 & 94.6 & 90.4 & 60.7 & 88.6 & -- \\
			Co-learn$^\dagger$$_{L/14}$~\cite{25colearn} & \citeye{IJCV25} & SF & U+M & 98.9 & 93.2 & 81.0 & 83.0 & 98.6 & \textbf{98.8} & \textbf{95.7} & 84.8 & 94.8 & 97.3 & 95.1 & 41.6 & 88.6 & -- \\
			DTKI$^\dagger$$_{B/32}$~\cite{26dtki} & \citeye{IPM26} & SF & U+M & 97.7 & 87.7 & 87.5 & 82.7 & 97.3 & 98.3 & 93.3 & 85.1 & 95.3 & 96.3 & 94.0 & 73.9 & 90.8 & -- \\
			\cline{5-18}
			\rule{0pt}{10pt}CLIP-RN101~\cite{radford2021learning} & \citeye{ICML21} & SFF & M & 98.2 & 83.9 & 90.5 & 73.5 & 97.2 & 84.0 & 95.3 & 65.7 & 79.4 & 89.9 & 91.8 & 63.3 & 84.4 & -- \\
			ImCapSFDA~\cite{25imcapda} & \citeye{ESWA26} & SFF & M & \textbf{99.4} & 91.0 & \textbf{93.6} & \textbf{79.0} & \textbf{99.3} & \textbf{94.3} & \textbf{96.2} & 54.4 & \textbf{94.5} & \textbf{99.5} & \textbf{94.7} & 68.5 & 88.7 $\pm$ 0.18 & 88.7 $\pm$ 0.18 \\
			\rowcolor{gray!25}DSSG & \citeye{Ours} & SFF & M & 99.3 & \textbf{91.6} & 93.1 & 77.8 & 99.2 & 93.8 & 95.7 & \textbf{69.7} & 92.2 & \textbf{99.5} & 94.4 & 69.8 & 89.7 $\pm$ 0.10 & 89.8 $\pm$ 0.02 \\
			\rowcolor{gray!25}DSSG-PAC & \citeye{Ours} & SFF & M & \textbf{99.4} & \textbf{91.6} & 93.0 & 77.9 & 99.2 & 94.1 & 95.7 & 69.2 & 92.5 & \textbf{99.5} & 94.4 & \textbf{69.9} & \textbf{89.7} $\pm$ 0.17 & \textbf{89.8} $\pm$ 0.10 \\
			\hline
			\hline
			\multicolumn{1}{c}{\textit{ViT-B/16:}} & \\
			CLIP-ViT/B-16~\cite{radford2021learning} & \citeye{ICML21} & SFF & M & 99.3 & 91.7 & \textbf{93.9} & 74.3 & 98.4 & 94.3 & 90.3 & 78.2 & 78.3 & 97.3 & 95.2 & 64.8 & 88.0 & -- \\
			ImCapSFDA~\cite{25imcapda} & \citeye{ESWA26} & SFF & M & \textbf{99.8} & \textbf{93.4} & 93.0 & 78.2 & \textbf{99.6} & \textbf{98.8} & \textbf{97.3} & 79.1 & \textbf{94.4} & \textbf{98.9} & 96.3 & \textbf{73.0} & 91.8 $\pm$ 0.07 & 91.8 $\pm$ 0.06 \\
			\rowcolor{gray!25}DSSG & \citeye{Ours} & SFF & M & 99.7 & 93.2 & 92.4 & \textbf{81.3} & 99.5 & 98.7 & 97.0 & \textbf{82.5} & 93.0 & 98.7 & \textbf{96.7} & 71.5 & \textbf{92.0} $\pm$ 0.03 & \textbf{92.0} $\pm$ 0.03 \\
			\rowcolor{gray!25}DSSG-PAC & \citeye{Ours} & SFF & M & 99.7 & 93.3 & 92.4 & \textbf{81.3} & 99.5 & \textbf{98.8} & 97.0 & 82.4 & 93.1 & 98.6 & \textbf{96.7} & 71.4 & \textbf{92.0} $\pm$ 0.04 & \textbf{92.0} $\pm$ 0.04 \\
			\hline
		\end{tabular}
	\end{adjustbox}
\end{table*}

\begin{table*}[t]
	\caption{Component ablation of DSSG under strict SFF-DA. Domain-wise results are reported as mean last-checkpoint accuracy (\%) $\pm$ standard deviation. Parentheses in Avg. and VisDA denote changes from DSSG in percentage points.}
	\label{Tab-sfda-ab-dssg-without}
	\centering
	\begin{adjustbox}{width=\linewidth,center}
		\begin{tabular}{@{}lccclccccll@{}}
			\toprule
			& \multicolumn{9}{c}{CLIP-RN50} & \multicolumn{1}{c}{CLIP-RN101} \\
			\cmidrule(lr){2-10}\cmidrule(lr){11-11}
			\multirow{2}{*}{Variant} & \multicolumn{4}{c}{Office-31 (31 classes)} & \multicolumn{5}{c}{Office-Home (65 classes)} & \multicolumn{1}{c}{VisDA (12 classes)} \\
			\cmidrule(lr){2-5}\cmidrule(lr){6-10}\cmidrule(lr){11-11}
			& A & D & W & \multicolumn{1}{c}{Avg.} & A & C & P & R & \multicolumn{1}{c}{Avg.} & \multicolumn{1}{c}{R} \\
			\midrule
			\rowcolor{gray!25} \textbf{DSSG} & 83.93~$\pm$~0.41 & 92.97~$\pm$~1.25 & 93.96~$\pm$~0.38 & 90.29 & 84.52~$\pm$~0.20 & 83.67~$\pm$~0.31 & 94.50~$\pm$~0.20 & 92.44~$\pm$~0.11 & 88.78 & 89.67~$\pm$~0.10 \\
			w/o DSG & 80.84~$\pm$~0.24 & 86.88~$\pm$~0.23 & 91.53~$\pm$~1.72 & 86.42~\textcolor{blue!50!black}{(-3.87)} & 83.00~$\pm$~0.46 & 80.61~$\pm$~0.17 & 92.00~$\pm$~0.08 & 91.74~$\pm$~0.00 & 86.84~\textcolor{blue!50!black}{(-1.94)} & 88.79~$\pm$~0.08~\textcolor{blue!50!black}{(-0.88)} \\
			w/o $\mathcal{L}_{\mathrm{cmkd}}^*$ & 81.35~$\pm$~0.16 & 89.49~$\pm$~0.12 & 91.74~$\pm$~0.07 & 87.53~\textcolor{blue!50!black}{(-2.76)} & 82.45~$\pm$~0.37 & 82.23~$\pm$~0.23 & 92.02~$\pm$~0.25 & 91.40~$\pm$~0.13 & 87.02~\textcolor{blue!50!black}{(-1.76)} & 89.68~$\pm$~0.18~\textcolor{red!75!black}{(+0.01)} \\
			w/o $\mathcal{L}_{\mathrm{con}}$ & 75.97~$\pm$~3.59 & 76.31~$\pm$~1.57 & 74.72~$\pm$~1.53 & 75.66~\textcolor{blue!50!black}{(-14.63)} & 69.34~$\pm$~1.73 & 41.64~$\pm$~4.63 & 89.75~$\pm$~0.30 & 88.51~$\pm$~0.30 & 72.31~\textcolor{blue!50!black}{(-16.47)} & 10.35~$\pm$~0.23~\textcolor{blue!50!black}{(-79.32)} \\
			w/o $\mathcal{L}_{\mathrm{st}}$ & 78.63~$\pm$~0.22 & 89.69~$\pm$~0.50 & 89.01~$\pm$~1.74 & 85.78~\textcolor{blue!50!black}{(-4.51)} & 60.54~$\pm$~1.34 & 55.24~$\pm$~1.69 & 71.11~$\pm$~1.21 & 70.48~$\pm$~4.98 & 64.34~\textcolor{blue!50!black}{(-24.44)} & 70.34~$\pm$~3.41~\textcolor{blue!50!black}{(-19.33)} \\
			\midrule
			& \multicolumn{10}{c}{CLIP-ViT-B/16} \\
			\cmidrule(lr){2-11}
			\multirow{2}{*}{Variant} & \multicolumn{4}{c}{Office-31 (31 classes)} & \multicolumn{5}{c}{Office-Home (65 classes)} & \multicolumn{1}{c}{VisDA (12 classes)} \\
			\cmidrule(lr){2-5}\cmidrule(lr){6-10}\cmidrule(lr){11-11}
			& A & D & W & \multicolumn{1}{c}{Avg.} & A & C & P & R & \multicolumn{1}{c}{Avg.} & \multicolumn{1}{c}{R} \\
			\midrule
			\rowcolor{gray!25} \textbf{DSSG} & 87.55~$\pm$~0.13 & 98.39~$\pm$~0.00 & 97.23~$\pm$~0.00 & 94.39 & 90.03~$\pm$~0.04 & 89.20~$\pm$~0.15 & 97.31~$\pm$~0.05 & 94.64~$\pm$~0.09 & 92.79 & 92.01~$\pm$~0.03 \\
			w/o DSG & 85.41~$\pm$~0.07 & 91.77~$\pm$~0.69 & 92.58~$\pm$~0.21 & 89.92~\textcolor{blue!50!black}{(-4.47)} & 88.37~$\pm$~0.06 & 88.08~$\pm$~0.08 & 94.46~$\pm$~0.15 & 94.24~$\pm$~0.02 & 91.29~\textcolor{blue!50!black}{(-1.50)} & 91.79~$\pm$~0.08~\textcolor{blue!50!black}{(-0.22)} \\
			w/o $\mathcal{L}_{\mathrm{cmkd}}^*$ & 85.00~$\pm$~0.13 & 91.77~$\pm$~0.20 & 95.22~$\pm$~0.00 & 90.66~\textcolor{blue!50!black}{(-3.73)} & 89.36~$\pm$~0.09 & 88.05~$\pm$~0.04 & 96.86~$\pm$~0.01 & 94.38~$\pm$~0.07 & 92.16~\textcolor{blue!50!black}{(-0.63)} & 92.00~$\pm$~0.02~\textcolor{blue!50!black}{(-0.01)} \\
			w/o $\mathcal{L}_{\mathrm{con}}$ & 85.50~$\pm$~0.29 & 87.48~$\pm$~0.42 & 89.77~$\pm$~0.07 & 87.59~\textcolor{blue!50!black}{(-6.80)} & 86.72~$\pm$~0.28 & 82.01~$\pm$~0.85 & 94.17~$\pm$~0.05 & 94.15~$\pm$~0.09 & 89.26~\textcolor{blue!50!black}{(-3.53)} & 90.74~$\pm$~0.06~\textcolor{blue!50!black}{(-1.27)} \\
			w/o $\mathcal{L}_{\mathrm{st}}$ & 84.88~$\pm$~0.07 & 92.30~$\pm$~0.50 & 92.32~$\pm$~0.65 & 89.84~\textcolor{blue!50!black}{(-4.55)} & 62.01~$\pm$~0.39 & 63.96~$\pm$~0.92 & 71.99~$\pm$~0.57 & 69.79~$\pm$~0.92 & 66.93~\textcolor{blue!50!black}{(-25.86)} & 86.84~$\pm$~0.29~\textcolor{blue!50!black}{(-5.17)} \\
			\bottomrule
		\end{tabular}
	\end{adjustbox}
\end{table*}

\subsection{Experimental Results}

We compare DSSG and DSSG-PAC with representative baselines from unimodal (U), unimodal-student-with-multimodal-teacher (U+M), and directly fine-tuned multimodal (M) settings.

\myparagraph{Results on Standard Benchmarks: Office-Home and Office-31.} 
As shown in Table~\ref{Tab-sfda-oh}, DSSG achieves \textbf{88.8\%} and \textbf{92.8\%} Avg. Last accuracy on Office-Home with ResNet-50 and ViT-B/16, outperforming ImCapSFDA by \textbf{4.6} and \textbf{2.1} percentage points, respectively. DSSG-PAC closely matches DSSG across both backbones. As shown in Table~\ref{Tab-sfda-o31}, DSSG achieves \textbf{90.3\%} and \textbf{94.4\%} Avg. Last accuracy on Office-31 with ResNet-50 and ViT-B/16, respectively, consistently outperforming existing SFF baselines, while DSSG-PAC maintains competitive performance across both backbones.

\myparagraph{Results on Large-Scale Benchmarks: Mini-DomainNet and VisDA.} Table~\ref{Tab-sfda-mini} and Table~\ref{Tab-sfda-visda-per} show that DSSG achieves \textbf{87.1\%} and \textbf{89.67\%} Avg. Last accuracy with ResNet, and \textbf{91.1\%} and \textbf{92.01\%} with ViT-B/16 on Mini-DomainNet and VisDA, respectively. DSSG-PAC closely matches DSSG on both large-scale benchmarks.

\myparagraph{Trade-offs and Fair Comparison: U+M vs. M Paradigms.}
Across the reported comparisons, DSSG consistently outperforms U+M baselines on both ResNet and ViT-B/16 backbones. These results indicate that U+M methods rely on source-trained initialization and external teacher guidance, whereas DSSG directly adapts a single VLM under the strict SFF setting without access to source models or external teachers. DSSG-PAC preserves this advantage while reducing the overhead of iteration-wise class-prompt encoding. 

\subsection{Ablation and Mechanism Analysis: DSSG and DSSG-PAC}

\myparagraph{Component Ablation of DSSG.} 
To provide a stricter evaluation, we re-evaluate DSSG under the SFF-DA setting via component-wise removal with last-checkpoint accuracy, complementing the incremental best-checkpoint ablation in Appendix~\ref{app:dssg-ab-best}. Removing DSG consistently degrades performance, while removing Dynamic CMKD generally results in lower accuracy, validating the complementary roles of dual-stream guidance and dynamic distillation. Consistent with prior findings on catastrophic forgetting during direct CLIP fine-tuning~\cite{2024cmkd}, removing $\mathcal{L}_{con}$ causes substantial performance degradation, including a severe drop from \textbf{89.67\%} to \textbf{10.35\%} on VisDA with RN101. By enforcing instance-level, fine-grained image--caption alignment, $\mathcal{L}_{con}$ guides both encoders toward target-specific semantics while preserving the pretrained cross-modal representation space, thereby mitigating catastrophic forgetting. Removing $\mathcal{L}_{st}$ also causes substantial degradation, highlighting its importance in converting dynamic teacher guidance into task-discriminative supervision. 

\noindent \textbf{Ablation Study on DSG Module.}
To examine the roles and complementarity of the two semantic streams in DSG, we remove $\mathcal{L}_{\mathrm{cmkd}}^*$ to exclude the effect of Dynamic CMKD. Without semantic guidance, direct CLIP fine-tuning suffers severe degradation with ResNet, indicating catastrophic forgetting, while ViT-B/16 does not exhibit comparable collapse. The anchor stream shows similar degradation with ResNet, indicating that category-level semantics without instance-specific target information are insufficient for stable adaptation. Adding the class-anchor stream further enhances overall performance, demonstrating the complementary roles of the two streams: captions provide target-specific plasticity, whereas class anchors preserve task-relevant categorical stability.

\begin{table}[t]
	\centering
	\caption{Ablation study of the DSG module. $\ddagger$ indicates catastrophic forgetting.}
	\label{tab:dsg-ablation}
	\begin{tabular}{lccccc}
		\toprule
		Method & O-31 & O-Home & MiniDN & VisDA & Avg. \\
		\midrule
		CLIP-RN & 71.90 & 72.10 & 72.50 & 84.40 & 75.23 \\
		FT w/o Stream$^\ddagger$ & 35.74 & 8.00 & 6.14 & 10.87 & 15.19 \\
		Anchor Stream$^\ddagger$ & 27.92 & 10.70 & 2.98 & 10.62 & 13.06 \\
		Caption Stream & 83.99 & 84.18 & \textbf{85.97} & 88.71 & 85.71 \\
		\textbf{Dual Stream (DSG)} & \textbf{87.53} & \textbf{87.02} & 85.91 & \textbf{89.68} & \textbf{87.54} \\
		\midrule
		CLIP-ViT-B/16 & 78.60 & 83.40 & 84.70 & 88.00 & 83.68 \\
		FT w/o Stream & 84.84 & 85.68 & 88.20 & 90.58 & 87.33 \\
		Anchor Stream & 85.09 & 86.25 & 88.46 & 90.62 & 87.61 \\
		Caption Stream & 89.53 & 90.69 & 90.78 & 91.80 & 90.70 \\
		\textbf{Dual Stream (DSG)} & \textbf{90.66} & \textbf{92.16} & \textbf{90.86} & \textbf{92.00} & \textbf{91.42} \\
		\bottomrule
	\end{tabular}
\end{table}

 \begin{figure*}[t]    
	\centering
	\begin{adjustbox}{width=1.0\linewidth,center}
		\includegraphics{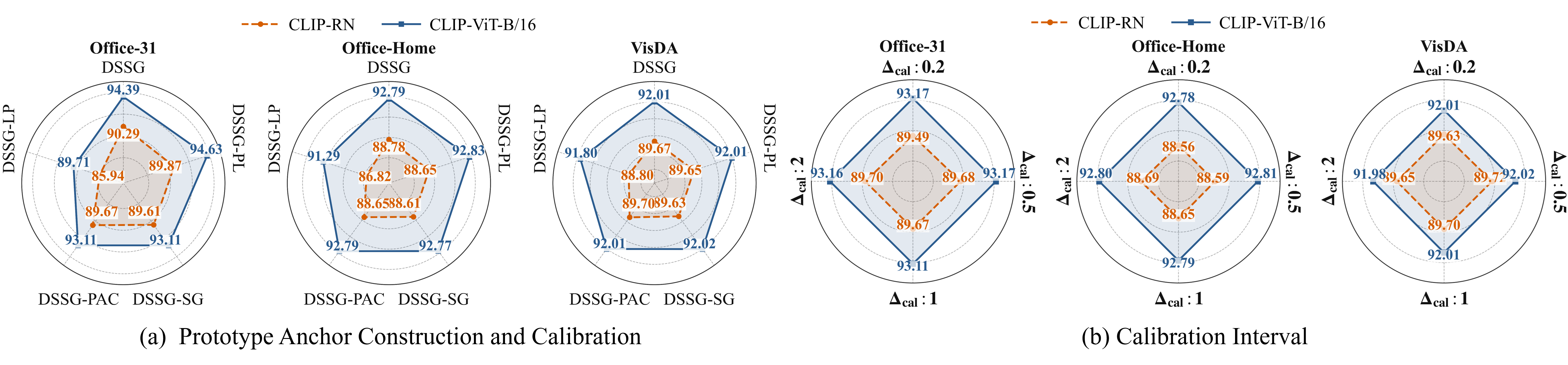}
	\end{adjustbox} 
	\caption{Prototype anchor analysis of DSSG-PAC: (a) construction and calibration; (b) sensitivity to calibration interval $\Delta_{\mathrm{cal}}$.}
	\label{fig:pac-ab}

\end{figure*}

\myparagraph{Analysis of Prototype Anchor Construction and Calibration.}
DSSG-PAC introduces two changes to the class-anchor stream: gradient detachment and periodic prototype calibration. We introduce DSSG-StopGrad to disentangle these two factors by detaching class-anchor gradients while retaining iteration-wise recomputation with the current $E_T$. DSSG-StopGrad exhibits noticeable degradation relative to DSSG mainly on Office-31, while remaining comparable on Office-Home and VisDA, suggesting that class-anchor gradients have limited overall impact. This degradation is primarily observed in target domains with fewer samples, possibly due to their greater sensitivity to gradient variations. In contrast, DSSG-PAC performs comparably to DSSG-StopGrad across all settings, indicating that periodically calibrated prototype anchors provide sufficient class guidance in most settings without requiring iteration-wise recomputation or gradient propagation through the anchor branch.

To further examine class-anchor construction, DSSG-PL adopts learnable context embeddings while retaining iteration-wise text encoding, whereas DSSG-LearnProto directly learns class prototypes without $E_T$. DSSG-PL provides no consistent gains across datasets or backbones, indicating limited benefit from prompt learning once the text encoder has been adapted through caption guidance, while retaining the same encoding overhead. In contrast, DSSG-LearnProto consistently degrades performance, showing that unconstrained prototypes cannot replace text-encoder-conditioned category semantics. Additionally, other PAC variants based on momentum and residual correction are evaluated in Appendix~\ref{app:pac}.

\begin{figure}[t]    
	
	\centering
	\begin{adjustbox}{width=1.0\linewidth,center}
		\includegraphics{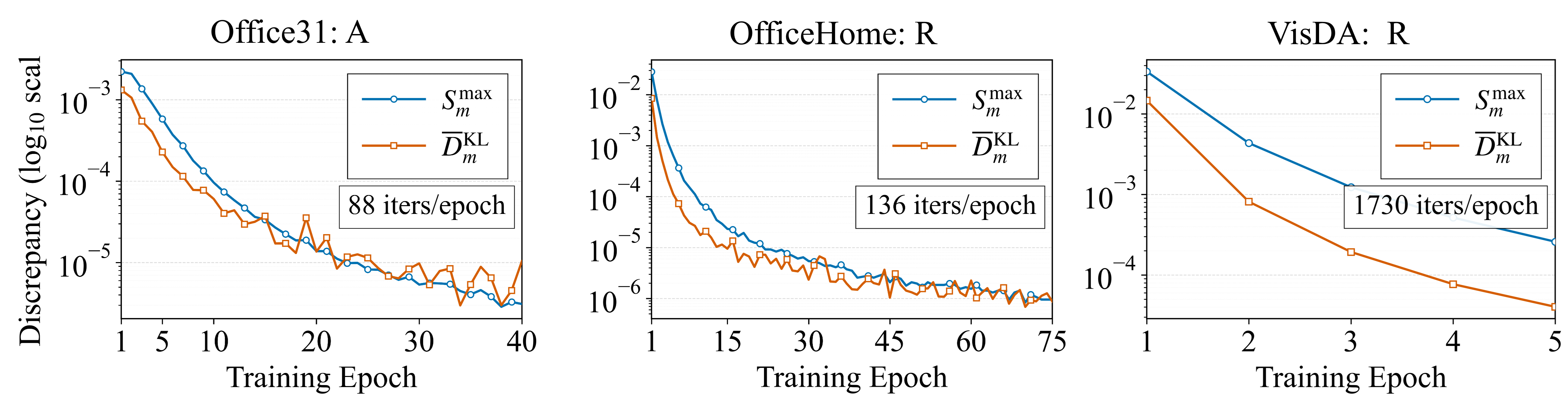}
	\end{adjustbox} 
	\caption{Prototype staleness and teacher prediction discrepancy.}
	\label{fig:pac-pskl}

\end{figure}

\myparagraph{Mechanism Analysis: Prototype Staleness and Teacher Prediction Consistency.}  
We investigate whether the periodically reused prototype anchors $g_{\mathrm{pac}}^{(m)}$ deviate from the class anchors $g_{\mathrm{anc}}^{(t)}$ generated by the evolving text encoder, and whether the resulting anchor discrepancy affects teacher predictions. For $t\in[t_m,t_{m+1})$, we quantify this anchor discrepancy as prototype staleness using their mean cosine distance, and record its maximum within each calibration interval:
\begin{equation}
	\begin{aligned}
		S_t
		&=
		1-\frac{1}{K}\sum_{k=1}^{K}
		\left\langle
		g_{\mathrm{anc},k}^{(t)},
		g_{\mathrm{pac},k}^{(m)}
		\right\rangle,
		\\
		S_m^{\max}
		&=
		\max_{t_m\leq t<t_{m+1}} S_t .
	\end{aligned}
\end{equation}
To assess teacher prediction consistency, we compute the batch-averaged KL divergence between the teacher predictions induced by $g_{\mathrm{anc}}^{(t)}$ and $g_{\mathrm{pac}}^{(m)}$, together with its mean over each calibration interval:
\begin{equation}
	\begin{aligned}
		D_t^{\mathrm{KL}}
		&=
		\frac{1}{B_t}
		\sum_{i=1}^{B_t}
		D_{\mathrm{KL}}
		\left(
		P_{i,t}^{\mathrm{tea,anc}}
		\parallel
		P_{i,t}^{\mathrm{tea,pac}}
		\right),
		\\
		\overline{D}_m^{\mathrm{KL}}
		&=
		\frac{1}{t_{m+1}-t_m}
		\sum_{t=t_m}^{t_{m+1}-1}
		D_t^{\mathrm{KL}} .
	\end{aligned}
\end{equation}
Within each calibration interval, prototype staleness starts from zero and increases approximately monotonically. As shown in Fig.~\ref{fig:pac-pskl}, maximum staleness and mean teacher KL rapidly decrease to below $10^{-4}$ and approximately $10^{-5}$,
respectively, demonstrating the rapid stabilization of class prototypes after early adaptation. Thus, periodically calibrated prototype anchors maintain teacher prediction consistency without requiring iteration-wise class-prompt encoding. 

\myparagraph{Effect of the Calibration Interval.}
As shown in Fig.~\ref{fig:pac-ab}, $\Delta_{\mathrm{cal}}=0.2, 0.5, 1,$ and $2$ yield comparable results, indicating that PAC is robust to variations in the calibration interval. This finding is consistent with the observed stabilization of prototype
anchors after early adaptation. Considering the calibration overhead and VisDA's five-epoch training schedule, we adopt
$\Delta_{\mathrm{cal}}=1$ as the default.

\subsection{Computational Efficiency Analysis}

\myparagraph{Training and Inference Efficiency Analysis.}
Table~\ref{Tab-model-efficiency-appendix} shows that DSSG adds no trainable parameters over ImCapSFDA but incurs adaptation overhead from iteration-wise class-prompt encoding. Periodic prototype calibration substantially reduces this cost in DSSG-PAC. At inference, DSSG computes class anchors only once, while DSSG-PAC directly uses the calibrated prototype anchors, leaving no additional downstream inference overhead. The cost of local caption generation is also reported, with batching substantially reducing the per-image latency.

\begin{table}[h]
	\caption{Computational efficiency comparison of DSSG and DSSG-PAC.}
	\label{Tab-model-efficiency-appendix}
	\centering
	\begin{adjustbox}{width=1.0\linewidth,center}
		\begin{tabular}{l|ccc|ccc}
			\toprule
			& \multicolumn{3}{c|}{\textbf{Training}} & \multicolumn{3}{c}{\textbf{Inference}} \\ 
			\cmidrule(lr){2-4} \cmidrule(lr){5-7}
			Model & \makecell{Trainable Params\\(M)} & \makecell{Iter Speed\\(it/s)} & \makecell{GPU Mem\\(MB)} & \makecell{Latency\\(ms)} & GFLOPs & \makecell{GPU Mem\\(MB)} \\ 
			\midrule
			CLIP-ViT-B/16 & 0.00 & - & - & 6.59 & 33.71 & 1228 \\
			ImCapSFDA & 124.09 & 12.93 & 5356 & 6.63 & 39.53 & 1228 \\
			DSSG & 124.09 & 12.09 & 5838 & 6.64 & 39.53 & 1228 \\  
			DSSG-PAC & 124.09 & 12.91 & 5570 & 6.63 & 39.53 & 1228 \\ 
			\midrule
	
			\makecell{Local Captioner\\(e.g., BLIP-3)} & \multicolumn{3}{c|}{\makecell{Caption Generation\\848 ($B$=1) / 203 ($B$=12) ms/img}} & \multicolumn{3}{c}{\makecell{GPU Memory\\\textasciitilde9842 MB}} \\

			\bottomrule
		\end{tabular}
	\end{adjustbox} 
	
\end{table}

\myparagraph{Adaptation-Time Efficiency Analysis.}
As shown in Table~\ref{Tab-sfda-training-time}, DSSG-PL moderately reduces adaptation time by avoiding repeated prompt-to-token-embedding conversion, whereas DSSG-PAC achieves an average 18.9\% reduction over DSSG and a runtime comparable to ImCapSFDA, further validating the efficiency of periodic prototype calibration.

\begin{table}[t]
	\caption{Total adaptation time (min) using CLIP-ViT-B/16, including both training and evaluation.}
	\label{Tab-sfda-training-time}
	\centering
	\begin{adjustbox}{width=\linewidth,center}
		\begin{tabular}{@{}lccccccccc@{}}
			\toprule
			& \multicolumn{9}{c}{CLIP-ViT-B/16} \\
			\cmidrule(lr){2-10}
			\multirow{2}{*}{Method} & \multicolumn{3}{c}{Office-31} & \multicolumn{4}{c}{Office-Home} & \multicolumn{1}{c}{VisDA} & \multicolumn{1}{c}{Overall} \\
			\cmidrule(lr){2-4}\cmidrule(lr){5-8}\cmidrule(lr){9-9}\cmidrule(lr){10-10}
			& A & D & W & A & C & P & R & Val & Reduction (\%)  \\
			\midrule
			ImCapSFDA & 5.5 & 3.8 & 4.7 & 8.9 & 15.5 & 15.7 & 16.4 & 18.9 & 19.1 \\
			DSSG & 6.5 & 4.2 & 5.8 & 11.8 & 20.5 & 20.8 & 21.4 & 21.4 & -- \\
			DSSG-PL & 6.0 & 4.1 & 5.5 & 11.0 & 19.2 & 19.5 & 20.2 & 20.3 & 5.7 \\
			\rowcolor{gray!25} DSSG-PAC & 5.4 & 3.8 & 5.0 & 8.9 & 15.3 & 15.5 & 16.3 & 18.9 & 18.9 \\
			\bottomrule
		\end{tabular}
	\end{adjustbox}
\end{table}

\subsection{Quantitative and Qualitative Analysis of Caption Semantics}

\myparagraph{Task-Relevant Proxies for the Semantic Quality of Captions.} 
Without target labels, the quality of generated captions and their impact on classification cannot be directly assessed, motivating the use of task-relevant proxy metrics. Explicit and implicit class matching~\cite{25imcapda} quantify ground-truth class-name and synonym occurrences, respectively, while class ambiguity measures captions matching multiple task classes. We further match each caption to class texts by cosine similarity and use the most similar class as the prediction to assess class discriminability.  

Table~\ref{Tab-caption-matching} shows that the ground-truth class name is explicitly matched in only a subset of captions, while implicit matching captures more synonymous class expressions, yet 16\%--39\% of captions still match multiple task classes, revealing intra-task class ambiguity and making direct class-term matching unreliable for standalone classification. Caption-to-class similarity prediction outperforms CLIP zero-shot on several domains but remains below caption-driven adaptation, suggesting that captions are more effective as complementary visual-semantic guidance than for classification alone.

\begin{table}[h] 
	\caption{Caption quality proxies and classification performance (\%).}
	\label{Tab-caption-matching}
	\centering

	\begin{adjustbox}{width=\linewidth,center}
		\begin{tabular}{@{}lcccccccc@{}}
			\toprule
			\multirow{2}{*}{Metric / Method} & \multicolumn{3}{c}{Office-31} & \multicolumn{4}{c}{Office-Home} & VisDA \\
			\cmidrule(lr){2-4}\cmidrule(lr){5-8}\cmidrule(lr){9-9}
			& A & D & W & A & C & P & R & Val \\
			\midrule
			Explicit matching rate & 51.76 & 58.63 & 56.86 & 59.46 & 63.41 & 78.37 & 72.64 & 57.38 \\
			Implicit matching rate & 58.47 & 71.08 & 69.43 & 70.37 & 75.58 & 86.96 & 81.09 & 62.86 \\
			Implicit ambiguity rate & 16.05 & 20.48 & 20.25 & 22.58 & 33.86 & 38.66 & 32.64 & 19.78 \\
			\midrule
			Class-Name Similarity & 77.32 & 79.72 & 82.01 & 77.75 & 78.97 & 91.71 & 84.71 & 77.47 \\
			Prompted-Class Similarity & 77.32 & 83.53 & 83.02 & 78.66 & 79.93 & 93.20 & 87.26 & 79.74 \\
			\midrule
			CLIP-ViT-B/16 & 78.90 & 79.90 & 76.90 & 82.30 & 71.30 & 89.20 & 89.90 & 88.00 \\
			ImCapSFDA & 85.19 & 91.84 & 91.57 & 87.52 & 86.99 & 94.36 & 93.89 & 91.75 \\
			DSSG (Ours) & 87.55 & 98.39 & 97.23 & 90.03 & 89.20 & 97.31 & 94.64 & 92.03 \\
			\bottomrule
		\end{tabular}
	\end{adjustbox}
\end{table}

\begin{figure*}[t]
	\centering
	\includegraphics[width=1.0\linewidth]{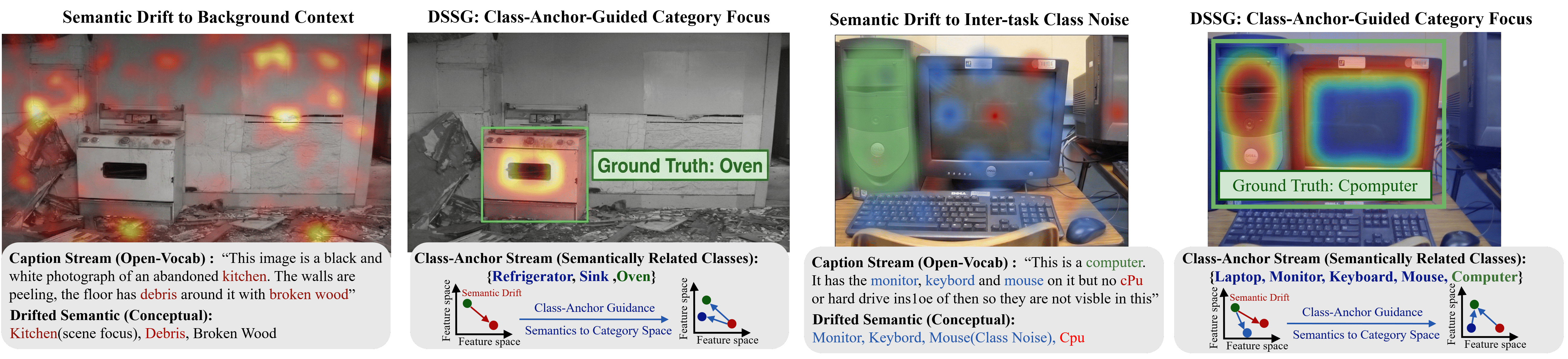}
	\caption{Caption-induced semantic drift and class-anchor guidance. Green, blue, and red mark ground truth, intra-task class noise, and background/OOD noise.}
	\label{fig:noisecaption} 

\end{figure*}

\myparagraph{Qualitative Analysis of Caption-Induced Semantic Drift.}
The quantitative results reveal class matching and ambiguity but not the specific distracting semantics in captions. As illustrated in Fig.~\ref{fig:noisecaption}, captions may mix class-discriminative cues with background/OOD semantics or cues from other task classes, shifting their semantic focus away from the target category. Thus, caption quality depends not only on descriptive richness but also on whether the semantic focus remains task-relevant. DSSG therefore complements open-vocabulary caption guidance with class-anchor guidance to preserve category-level focus during adaptation.

\begin{figure}[t]    
	
	\centering
	\begin{adjustbox}{width=1.0\linewidth,center}
		\includegraphics{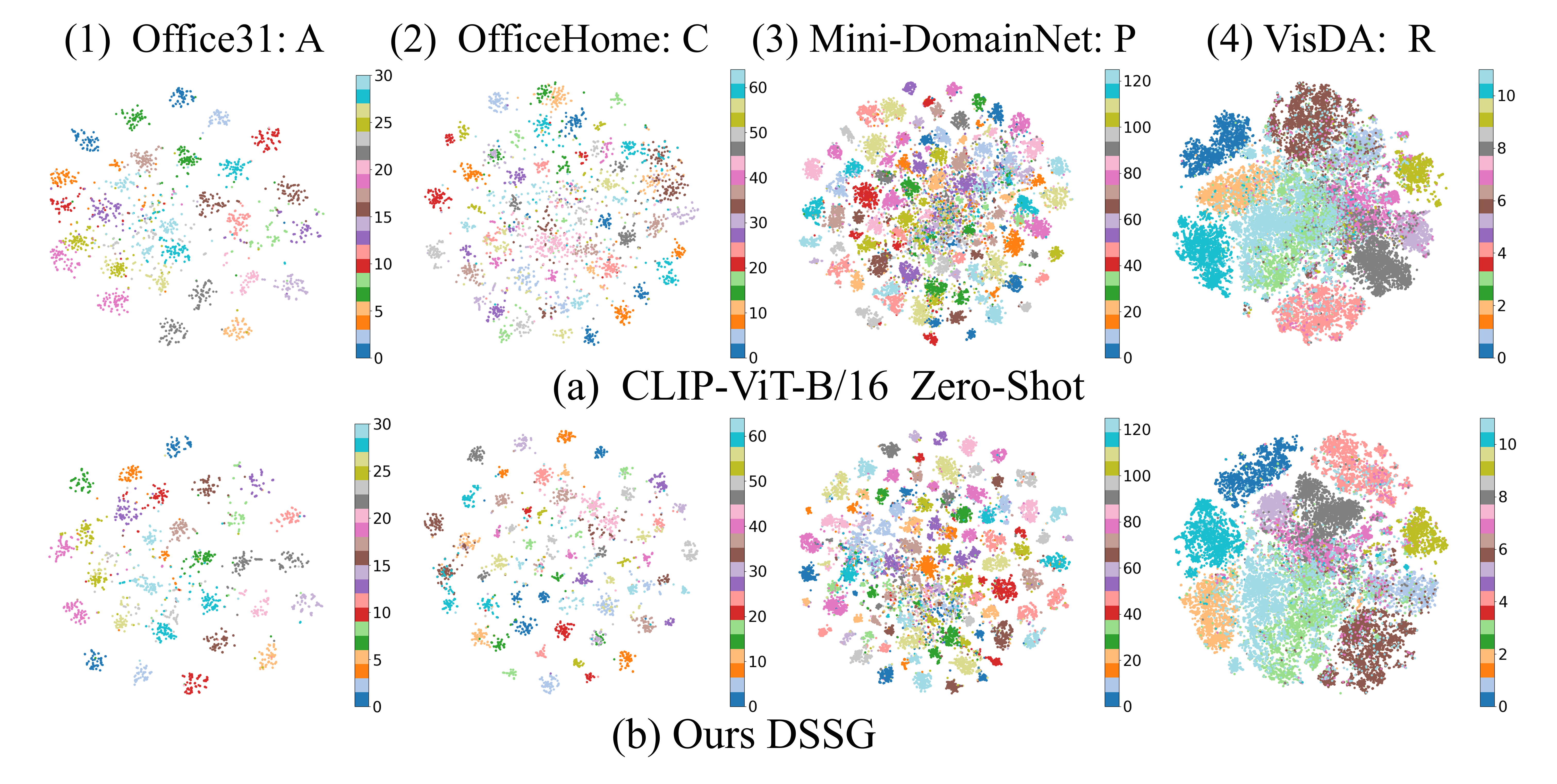}
	\end{adjustbox} 
	\caption{t-SNE visualizations across four benchmarks.}
	\label{Fig-tsne-vis}

\end{figure}

\myparagraph{Visualization Analysis.}  
t-SNE visualizations in Fig.~\ref{Fig-tsne-vis} show that DSSG yields more compact and better-separated target clusters than CLIP zero-shot across all four benchmarks, indicating improved target-domain discriminability.

\subsection{Discussion}  

\myparagraph{From Dual-Stream Semantic Guidance to Efficient Prototype Calibration.}
%Our analyses show that captions provide informative instance-level semantics but also introduce intra-task ambiguity and semantic drift. DSSG therefore uses dual-stream guidance to couple instance-level caption semantics with task-relevant class anchors, balancing adaptation and semantic stability. DSSG-PAC further improves efficiency through periodic calibration of prototype anchors while preserving adaptive category guidance.
Our analyses reveal that captions provide instance-specific semantics but may also introduce intra-task ambiguity and semantic drift. DSSG balances semantic plasticity and stability by coupling caption guidance with task-relevant class anchors, while Dynamic CMKD promotes teacher--student consistency. As the anchor space stabilizes during adaptation, DSSG-PAC periodically calibrates and reuses prototype anchors, reducing redundant text encoding while largely retaining the adaptive category guidance of DSSG.

\myparagraph{Strict Source-Fully-Free Adaptation.}
%DSSG further demonstrates the viability of strict SFF-DA without the task-specific source models required by U methods or the additional multimodal teachers used by U+M methods, while outperforming representative U and M methods.
DSSG operates under the strict SFF-DA protocol, requiring neither source data nor a task-specific source model. Unlike U methods that inherit source-supervised models and U+M methods that additionally employ external multimodal teachers, DSSG adapts a single pretrained VLM using only unlabeled target data and generated captions. Its competitive or superior results across these protocols demonstrate that generic vision--language knowledge can effectively compensate for unavailable source supervision.

\myparagraph{Comparison with Zero-Shot MLLM Inference.} 
%Table~\ref{tab:res-cap-sfda} compares DSSG with zero-shot MLLM results from~\cite{chen2024empowering}, including LLaVA-v1.6~\cite{liu2023visual}, InstructBLIP~\cite{dai2023instructblip}, and ShareGPT4V~\cite{chen2024sharegpt4v}. The MLLMs perform zero-shot classification through a VQA formulation with the target image and all task class names, followed by STS-based output matching using \textit{all-MiniLM-L6-v2} ($\sim$0.02B). On average, DSSG outperforms all three zero-shot MLLMs on Office-Home and MiniDomainNet and matches LLaVA-v1.6-34B on VisDA, while avoiding the substantial inference overhead of direct large-MLLM prediction. DSSG instead uses a lightweight CLIP backbone (0.12B) to fuse cross-modally aligned image and caption features for prediction, while captions can be obtained from locally deployed captioners, such as BLIP-3 (4.1B)~\cite{xue2024xgen}, online APIs, or offline pre-computation.
Table~\ref{tab:res-cap-sfda} compares DSSG with the zero-shot MLLM results reported in~\cite{chen2024empowering}, including LLaVA-v1.6~\cite{liu2023visual}, InstructBLIP~\cite{dai2023instructblip}, and ShareGPT4V~\cite{chen2024sharegpt4v}. These methods perform VQA-based classification using the target image and all class names, followed by STS matching with \textit{all-MiniLM-L6-v2} ($\sim$0.02B). DSSG outperforms these MLLMs on average on Office-Home and MiniDomainNet and matches LLaVA-v1.6-34B on VisDA. Instead of repeatedly invoking a large MLLM for prediction, DSSG adapts a lightweight CLIP backbone (0.12B), while captions may be generated locally, obtained via APIs, or pre-computed offline.

\myparagraph{Limitations and Future Work.}
%DSSG inherits the reliance on caption generation from caption-based adaptation: locally deployed captioners and offline pre-computation incur additional computation and storage overhead, respectively, while online APIs introduce communication latency. Future work will explore higher-quality caption generation and refinement for more reliable semantic guidance.
DSSG remains dependent on caption quality and acquisition. Locally deployed captioners introduce computation, offline captions require storage, and online APIs incur latency and potential privacy concerns. Moreover, task-irrelevant or biased descriptions may still affect adaptation despite class-anchor regularization. Future work will explore uncertainty-aware caption refinement, lightweight captioning, and adaptive prototype calibration for more reliable and efficient semantic guidance.

\begin{table}[t]
	\caption{Comparison with zero-shot MLLM inference}
	\label{tab:res-cap-sfda}
	\centering
	\begin{adjustbox}{width=1.0\linewidth}
		\begin{tabular}{lccccccccccc}
			\toprule
			\multirow{2}{*}{Method} & \multicolumn{5}{c}{Office-Home} & \multicolumn{5}{c}{MiniDomainNet} & VisDA \\
			\cmidrule(lr){2-6}\cmidrule(lr){7-11}\cmidrule(lr){12-12}
			& A & C & P & R & Avg. & C & P & R & S & Avg. & Val \\
			\midrule
			LLaVA-v1.6-34B & 87.0 & 78.3 & 93.7 & 89.5 & 87.1 & 85.5 & 84.4 & 91.0 & 83.7 & 86.2 & 92.1 \\
			InstructBLIP-XXL-12B & 82.2 & 82.0 & 91.6 & 88.8 & 86.2 & 86.7 & 82.5 & 89.0 & 83.0 & 85.3 & 86.7 \\
			ShareGPT4V-13B & 83.2 & 66.7 & 85.8 & 84.8 & 80.1 & 79.9 & 79.7 & 87.9 & 79.2 & 81.7 & 90.4 \\
			DSSG-0.12B (BLIP-3, 4.1B) & 90.1 & 89.4 & 97.4 & 94.6 & 92.9 & 91.1 & 83.7 & 94.9 & 87.6 & 89.3 & 92.1 \\
			\bottomrule
		\end{tabular}
	\end{adjustbox}
\end{table}

\section{Conclusion}
%In this paper, we investigate Source-Fully-Free Domain Adaptation (SFF-DA) and identify Dual Semantic Drift arising from fixed class embeddings and unconstrained caption semantics. To address this challenge, we propose DSSG, which couples instance-specific caption semantics with task-relevant class anchors, while Dynamic CMKD enforces teacher--student consistency during adaptation. We further introduce DSSG with Prototype Anchor Calibration (DSSG-PAC), which periodically calibrates prototype anchors to reduce redundant text-side computation while preserving adaptive category guidance. We further establish SFF-DA risk bounds that relate student risk to semantic-teacher quality and teacher--student discrepancy. Extensive experiments demonstrate state-of-the-art performance across multiple benchmarks, while the PAC variant largely preserves adaptation accuracy with 20\% lower training time. Overall, our work provides an effective and efficient framework for source-fully-free adaptation of vision-language models. 

We investigate strict Source-Fully-Free Domain Adaptation and identify Dual Semantic Drift arising from fixed class embeddings and unconstrained caption semantics. DSSG addresses these complementary failure modes by coupling instance-specific captions with dynamically updated class anchors, while Dynamic CMKD enforces consensus-aware teacher--student consistency. DSSG-PAC further replaces iteration-wise class-prompt encoding with periodic prototype calibration. Our theoretical analysis relates the student risk to semantic-teacher quality and margin-normalized teacher--student discrepancy, clarifying the conditions for reliable adaptation. Experiments across multiple benchmarks demonstrate strong performance under the strict SFF-DA protocol, while DSSG-PAC largely preserves accuracy with 20\% lower training time. These results establish an effective accuracy--efficiency trade-off for source-independent adaptation of vision-language models.

\bibliographystyle{IEEEtran}
\bibliography{refs}

\section{Appendix Material} 
%
%This supplementary material includes the following contents: 
%\begin{itemize}
%
%	\item Experimental Configurations.  
%	\item Per-category Results on VisDA-2017.
%	\item Optimization of Baseline Hyper-parameters $\lambda_1$ and $\lambda_2$.
%	\item Sensitivity Analysis of Hyper-parameter $\lambda_3$.
%	\item Computational Efficiency Analysis.
%	\item Visualization of Activation Maps.
%	\item Limitations and Future Work.
%	
%\end{itemize}

\subsection{Experimental Configurations} 
\label{app:exp_config} 
Table~\ref{tab:hyperparams} summarizes the experimental configurations and hyperparameter settings for all benchmarks and backbones to facilitate reproducibility.
\begin{table}[h]
	\centering
	\caption{Training configurations under the strict SFF-DA setting.}
	\label{tab:hyperparams}
	\begin{adjustbox}{width=1.0\linewidth,center}
		\begin{tabular}{lcccc}
			\toprule
			Parameter & Office-31 & Office-Home & Mini-DomainNet & VisDA-2017 \\
			\midrule
			Common Settings & \multicolumn{4}{c}{\makecell[l]{CLIP-RN50 (RN101 on VisDA) / ViT-B/16, WIT-400M  \\  
					SGD ($m=0.9$, weight decay 5e-4), Batch size 32, $\gamma=0.9$ \\ FC-Head LR: $1000\times$ encoder LR; \\ Seeds: $\{2020,2026,2027\}$; PyTorch, RTX 4090D}} \\
			\midrule
			Target Samples & \makecell[c]{A: 2,817 \\ D: 498 \\ W: 795} & \makecell[c]{A: 2,427 \\ C: 4,365 \\ P: 4,439 \\ R: 4,357} & \makecell[c]{C: 18,523 \\ P: 30,042 \\ R: 69,622 \\ S: 24,147} & R: 55,388 \\
			Epochs & \makecell[c]{A: 40 \\ D/W: 120} & 75 & 10 & 5 \\
			\midrule
			LR (RN) & 3e-6 & 3e-6 & 6e-7 & 5e-8 \\
			LR (ViT) & 3e-6 & 3e-6 & 1.5e-6 & 3e-7 \\
			$(\lambda_1,\lambda_2,\lambda_3)$ & $(0.5,0.1,0.5)$ & $(1.0,0.05,0.2)$ & $(0.3,0.2,0.4)$ & $(1.0,1.0,0.1)$ \\
			\bottomrule
		\end{tabular}
	\end{adjustbox}
\end{table}

\subsection{Incremental Ablation of DSSG under Best-Checkpoint Evaluation}
\label{app:dssg-ab-best}

Table~\ref{Tab-sfda-ab-dssg} complements the main-text component-removal ablation with an incremental best-checkpoint analysis.  DSG (\cnum{2}) and static CMKD (\cnum{3}) independently improve the baseline and provide further gains when combined (\cnum{4}). Dynamic CMKD (\cnum{6}) consistently outperforms its static counterpart across datasets and backbones, while the degradation of \cnum{5} confirms the importance of caption-based contrastive alignment.

\begin{table*}[t]
	\caption{Ablation study of DSSG. ID~\Circled{1} denotes the ImCapSFDA baseline. DSG denotes the dual-stream guiding structure.}
	\label{Tab-sfda-ab-dssg}
	\centering
	\begin{adjustbox}{width=1.0\linewidth,center}
		\begin{tabular}{c@{\hskip 3pt}c@{\hskip 5pt}l@{\hskip 7pt}cccc ccccc ccccc c}
			\toprule
			& & & \multicolumn{14}{c}{CLIP-RN50} & RN101 \\
			\cmidrule(lr){4-17} \cmidrule(lr){18-18}
			\multirow{2}{*}{ID} & \multirow{2}{*}{DSG} & \multirow{2}{*}{Loss} & \multicolumn{4}{c}{Office31(31 classes)} & \multicolumn{5}{c}{Office-Home(65 classes)} & \multicolumn{5}{c}{MiniDomainNet(126 classes)} & VisDA(12 classes) \\
			\cmidrule(lr){4-7} \cmidrule(lr){8-12} \cmidrule(lr){13-17} \cmidrule(lr){18-18}
			& & & A & D & W & Avg. & A & C & P & R & Avg. & C & P & R & S & Avg. & Real \\
			\midrule
			\footnotesize{\Circled{1}} & - & $\mathcal{L}_{con}+\mathcal{L}_{st}$ & 78.13 & 88.55 & 86.04 & 84.24 & 81.87 & 77.25 & 88.92 & 89.95 & 84.50 & 83.17 & 77.62 & 90.95 & 78.89 & 82.66 & 88.95 \\
			\footnotesize{\Circled{2}} & \ding{51} & $\mathcal{L}_{con}+\mathcal{L}_{st}$ & 81.54 & 90.16 & 92.33 & 88.01 & 82.94 & 82.57 & 92.25 & 91.74 & 87.38 & 83.49 & 76.35 & 92.06 & 78.89 & 82.70 & 89.72 \\
			\footnotesize{\Circled{3}} & - & $\mathcal{L}_{con}+\mathcal{L}_{st}+\mathcal{L}_{cmkd}$ & 82.00 & 90.16 & 92.70 & 88.29 & 83.48 & 80.38 & 92.43 & 92.06 & 87.09 & \textbf{85.87} & 77.62 & 92.22 & 80.32 & 84.01 & 89.12 \\
			\footnotesize{\Circled{4}} & \ding{51} & $\mathcal{L}_{con}+\mathcal{L}_{st}+\mathcal{L}_{cmkd}$ & 83.95 & 91.77 & 93.58 & 89.77 & 85.25 & 83.96 & 94.30 & 92.66 & 89.04 & 84.92 & 76.35 & 92.70 & \textbf{80.48} & 83.61 & 89.60 \\
			\footnotesize{\Circled{5}} & \ding{51} & $\mathcal{L}_{st}+\mathcal{L}_{cmkd}^{*}$ & 84.06 & 81.12 & 82.64 & 83.35 & 83.07 & 69.19 & 91.24 & 87.80 & 82.83 & 84.76 & 76.67 & 91.75 & 76.51 & 82.42 & 84.07 \\
			\footnotesize{\Circled{6}} & \ding{51} & $\mathcal{L}_{con}+\mathcal{L}_{st}+\mathcal{L}_{cmkd}^{*}$ & \textbf{85.27} & \textbf{92.57} & \textbf{94.09} & \textbf{90.64} & \textbf{85.41} & \textbf{84.08} & \textbf{94.80} & \textbf{92.82} & \textbf{89.28} & 85.56 & \textbf{77.94} & \textbf{92.86} & 80.00 & \textbf{84.09} & \textbf{89.98} \\
			\midrule
			\multicolumn{3}{c}{} & \multicolumn{15}{c}{CLIP-ViT-B/16} \\
			\cmidrule(lr){4-18}
			\multirow{2}{*}{ID} & \multirow{2}{*}{DSG} & \multirow{2}{*}{Loss} & \multicolumn{4}{c}{Office31(31 classes)} & \multicolumn{5}{c}{Office-Home(65 classes)} & \multicolumn{5}{c}{MiniDomainNet(126 classes)} & VisDA(12 classes) \\
			\cmidrule(lr){4-7} \cmidrule(lr){8-12} \cmidrule(lr){13-17} \cmidrule(lr){18-18}
			& & & A & D & W & Avg. & A & C & P & R & Avg. & C & P & R & S & Avg. & Real \\
			\midrule
			\footnotesize{\Circled{1}} & - & $\mathcal{L}_{con}+\mathcal{L}_{st}$ & 85.20 & 92.17 & 93.08 & 90.15 & 87.76 & 87.06 & 94.50 & 94.29 & 90.90 & 90.16 & 83.33 & 94.29 & 86.83 & 88.65 & 91.75 \\
			\footnotesize{\Circled{2}} & \ding{51} & $\mathcal{L}_{con}+\mathcal{L}_{st}$ & 85.55 & 95.38 & 95.35 & 92.09 & 89.62 & 88.06 & 96.91 & 94.38 & 92.24 & 89.68 & 82.86 & \textbf{94.92} & 87.30 & 88.69 & 92.08 \\
			\footnotesize{\Circled{3}} & - & $\mathcal{L}_{con}+\mathcal{L}_{st}+\mathcal{L}_{cmkd}$ & 86.16 & 92.97 & 92.20 & 90.44 & 88.75 & 88.04 & 94.66 & 94.22 & 91.42 & 90.32 & 83.33 & 94.13 & 87.14 & 88.73 & 91.78 \\
			\footnotesize{\Circled{4}} & \ding{51} & $\mathcal{L}_{con}+\mathcal{L}_{st}+\mathcal{L}_{cmkd}$ & 86.01 & 96.18 & 94.97 & 92.39 & 90.07 & 89.21 & 97.25 & 94.58 & 92.78 & \textbf{91.11} & 83.17 & \textbf{94.92} & \textbf{87.62} & 89.21 & 92.07 \\
			\footnotesize{\Circled{5}} & \ding{51} & $\mathcal{L}_{st}+\mathcal{L}_{cmkd}^{*}$ & 87.50 & 87.95 & 91.19 & 88.88 & 87.80 & 82.96 & 94.32 & 94.15 & 89.81 & 90.16 & 81.75 & \textbf{94.92} & 86.19 & 88.26 & 90.68 \\
			\footnotesize{\Circled{6}} & \ding{51} & $\mathcal{L}_{con}+\mathcal{L}_{st}+\mathcal{L}_{cmkd}^{*}$ & \textbf{87.79} & \textbf{98.59} & \textbf{98.36} & \textbf{94.91} & \textbf{90.11} & \textbf{89.39} & \textbf{97.36} & \textbf{94.61} & \textbf{92.87} & \textbf{91.11} & \textbf{83.65} & \textbf{94.92} & \textbf{87.62} & \textbf{89.33} & \textbf{92.13} \\
			\bottomrule
		\end{tabular}
	\end{adjustbox}
\end{table*}

\subsection{Prototype-Anchor Calibration Strategy}
\label{app:pac}

Table~\ref{Tab-sfda-ab-pac-variants} compares DSSG-PAC with momentum- and residual-based calibration variants. At each periodic calibration event, DSSG-PAC updates the prototype cache with the current text-encoded class prototypes, i.e., $P_j^{\mathrm{cache}}\leftarrow P_j^{\mathrm{cur}}$, and keeps it fixed until the next calibration. PAC-Momentum introduces temporal smoothing as
\begin{equation}
	P_j^{\mathrm{cache}}\leftarrow\operatorname{Norm}\left(mP_{j-1}^{\mathrm{cache}}+(1-m)P_j^{\mathrm{cur}}\right),
\end{equation}
while the residual variants use
\begin{equation}
	\widetilde{P}_j=\operatorname{Norm}\left(P_j^{\mathrm{cache}}+\Delta P_j\right).
\end{equation}
PAC-Residual (Interval) resets $\Delta P_j$ after each calibration, whereas PAC-Residual (Cross) retains it across calibration intervals.

As shown in Table~\ref{Tab-sfda-ab-pac-variants}, strong momentum smoothing ($m=0.9$) degrades performance, while weak smoothing ($m=0.1$) and the residual variants provide only marginal and inconsistent changes relative to DSSG-PAC. These results indicate that the periodic calibration in DSSG-PAC is sufficient to track the evolving text representation space without additional smoothing or residual correction.

\begin{table*}[t]
	\caption{Ablation study on PAC variants. Results are reported as mean accuracy (\%) $\pm$ standard deviation.}
	\label{Tab-sfda-ab-pac-variants}
	\centering
	
	\begin{adjustbox}{width=\linewidth,center}
		\begin{tabular}{@{}lcccccccccc@{}}
			\toprule
%			& \multicolumn{9}{c}{CLIP-RN50} & \multicolumn{1}{c}{CLIP-RN101} \\
%			\cmidrule(lr){2-10}\cmidrule(lr){11-11}
%			\multirow{2}{*}{Variant} & \multicolumn{4}{c}{Office-31} & \multicolumn{5}{c}{Office-Home} & VisDA-2017 \\
%			\cmidrule(lr){2-5}\cmidrule(lr){6-10}\cmidrule(lr){11-11}
%			& A & D & W & Avg. & A & C & P & R & Avg. & Val \\
%			\midrule
%			
%			PAC-Direct & 83.51~$\pm$~0.32 & 92.03~$\pm$~1.23 & 93.46~$\pm$~0.82 & 89.67~$\pm$~0.49 & 84.52~$\pm$~0.35 & 83.51~$\pm$~0.27 & 94.20~$\pm$~0.11 & 92.39~$\pm$~0.15 & 88.65~$\pm$~0.09 & 89.70~$\pm$~0.17 \\
%			PAC-Momentum ($m = 0.9$) & 83.71~$\pm$~0.06 & 90.36~$\pm$~0.53 & 93.38~$\pm$~0.38 & 89.15~$\pm$~0.08 & 84.16~$\pm$~0.10 & 83.63~$\pm$~0.29 & 94.13~$\pm$~0.12 & 92.32~$\pm$~0.21 & 88.56~$\pm$~0.02 & 89.10~$\pm$~0.24 \\
%			PAC-Momentum ($m = 0.1$) & 83.66~$\pm$~0.38 & 91.56~$\pm$~1.41 & 93.04~$\pm$~0.59 & 89.42~$\pm$~0.70 & 84.20~$\pm$~0.17 & 83.53~$\pm$~0.21 & 94.11~$\pm$~0.11 & 92.40~$\pm$~0.11 & 88.56~$\pm$~0.13 & 89.64~$\pm$~0.11 \\
%			PAC-Residual (Interval) & 83.93~$\pm$~0.27 & 92.43~$\pm$~1.11 & 93.25~$\pm$~0.72 & 89.87~$\pm$~0.51 & 84.32~$\pm$~0.27 & 83.71~$\pm$~0.17 & 94.17~$\pm$~0.13 & 92.40~$\pm$~0.22 & 88.65~$\pm$~0.11 & 89.66~$\pm$~0.23 \\
%			PAC-Residual (Cross) & 83.50~$\pm$~0.27 & 91.63~$\pm$~2.13 & 93.08~$\pm$~1.85 & 89.41~$\pm$~0.82 & 84.12~$\pm$~0.12 & 83.76~$\pm$~0.04 & 94.26~$\pm$~0.06 & 92.45~$\pm$~0.14 & 88.65~$\pm$~0.02 & 89.64~$\pm$~0.12 \\
			
%			\midrule
%			& \multicolumn{10}{c}{CLIP-ViT-B/16} \\
%			\cmidrule(lr){2-11}
			\multirow{2}{*}{Variant} & \multicolumn{4}{c}{Office-31} & \multicolumn{5}{c}{Office-Home} & VisDA-2017 \\
			\cmidrule(lr){2-5}\cmidrule(lr){6-10}\cmidrule(lr){11-11}
			& A & D & W & Avg. & A & C & P & R & Avg. & R \\
			\midrule
			
			DSSG-PAC & 85.16~$\pm$~0.12 & 97.46~$\pm$~0.23 & 96.73~$\pm$~0.82 & 93.11~$\pm$~0.18 & 90.06~$\pm$~0.08 & 89.22~$\pm$~0.12 & 97.31~$\pm$~0.10 & 94.57~$\pm$~0.01 & 92.79~$\pm$~0.02 & 92.01~$\pm$~0.04 \\
			PAC-Momentum ($m = 0.9$) & 85.28~$\pm$~0.07 & 97.26~$\pm$~0.12 & 95.68~$\pm$~1.34 & 92.74~$\pm$~0.45 & 89.83~$\pm$~0.05 & 89.12~$\pm$~0.13 & 97.25~$\pm$~0.14 & 94.44~$\pm$~0.04 & 92.66~$\pm$~0.00 & 91.87~$\pm$~0.05 \\
			PAC-Momentum ($m = 0.1$) & 85.15~$\pm$~0.10 & 97.46~$\pm$~0.23 & 96.81~$\pm$~0.63 & 93.14~$\pm$~0.11 & 90.08~$\pm$~0.02 & 89.21~$\pm$~0.10 & 97.32~$\pm$~0.11 & 94.60~$\pm$~0.05 & 92.80~$\pm$~0.01 & 92.00~$\pm$~0.03 \\
			PAC-Residual (Interval) & 85.16~$\pm$~0.09 & 97.46~$\pm$~0.23 & 96.86~$\pm$~0.63 & 93.16~$\pm$~0.11 & 90.08~$\pm$~0.12 & 89.21~$\pm$~0.14 & 97.31~$\pm$~0.10 & 94.56~$\pm$~0.04 & 92.79~$\pm$~0.04 & 92.01~$\pm$~0.04 \\
			PAC-Residual (Cross) & 85.15~$\pm$~0.10 & 97.46~$\pm$~0.23 & 96.90~$\pm$~0.63 & 93.17~$\pm$~0.12 & 90.06~$\pm$~0.08 & 89.22~$\pm$~0.13 & 97.31~$\pm$~0.10 & 94.57~$\pm$~0.04 & 92.79~$\pm$~0.02 & 92.01~$\pm$~0.03 \\
			\bottomrule
		\end{tabular}
	\end{adjustbox}
\end{table*}

\subsection{Sensitivity Analysis of Hyper-parameter $\lambda_3$}
\label{appendix:lambda3}
\begin{figure*}[t]     
	\begin{adjustbox}{width=0.9\linewidth,center}
		\includegraphics{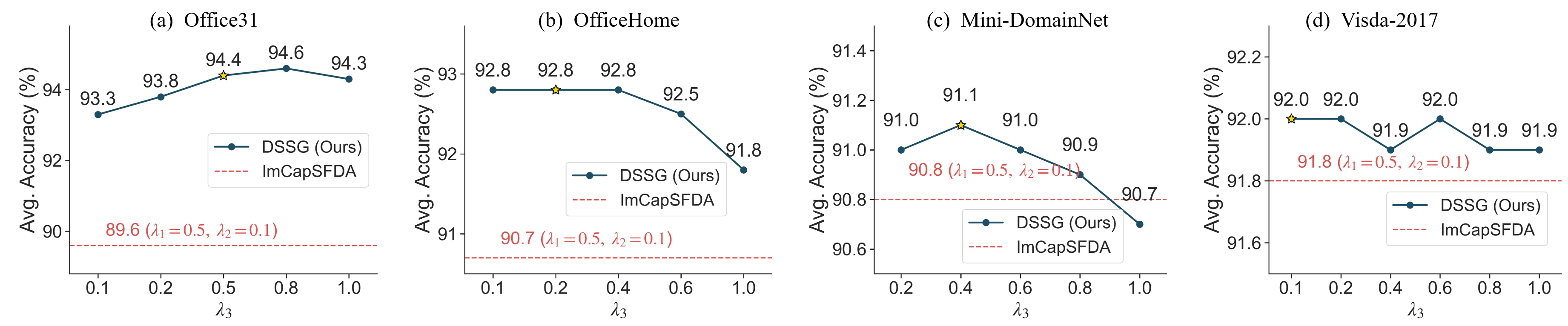}
	\end{adjustbox} 
	\caption{Hyper-Parameter Sensitivity Analysis of  $\lambda_3$  on ViT-B/16}
	\label{Fig-hyper-lambda3}   
	
\end{figure*}

To avoid task-specific hyperparameter tuning, we follow the default settings of ImCapSFDA for $\lambda_1$ and $\lambda_2$ without additional search. We then vary only the DSSG-specific $\lambda_3$ for sensitivity analysis. As shown in Fig.~\ref{Fig-hyper-lambda3}, DSSG maintains stable performance across a broad range of $\lambda_3$ values while consistently outperforming the ImCapSFDA baseline, indicating low sensitivity to $\lambda_3$.

\subsection{Proofs for the SFF-DA Risk Analysis}
\label{app:risk_proof}

This appendix proves Theorem~1 and Corollary~1 and states precisely the conditions under which the population certificate can be estimated from finite unlabeled data.

\subsubsection{Notation and tie convention}

For $p,q\in\Delta^{K-1}$, let $h_p(x)$ and $h_q(x)$ be top-1 classifiers under the same deterministic tie-breaking rule. Write $q_{(1)}(x)\ge q_{(2)}(x)$ for the two largest coordinates of $q(x)$ and $m_q(x)=q_{(1)}(x)-q_{(2)}(x)$. If $m_q(x)=0$, Eq.~\eqref{eq:certificate} assigns $\Psi(p,q;x)=1$; hence tied teacher predictions are conservatively treated as uncertified.

\subsubsection{A pointwise margin--discrepancy certificate}

\noindent\textbf{Lemma 1.}
For every $x$,
\begin{equation}
	\mathbf 1[h_p(x)\neq h_q(x)]\leq \Psi(p,q;x).
	\label{eq:pointwise_certificate}
\end{equation}

\noindent\emph{Proof.}
If $h_p(x)=h_q(x)$, the left-hand side is zero and the claim is immediate. Otherwise, let $a=h_q(x)$ and $j=h_p(x)\neq a$. Since $j$ is selected by the student, $p_j\ge p_a$. Since $a$ is selected by the teacher,
\begin{equation}
	q_a-q_j\ge q_{(1)}-q_{(2)}=m_q.
\end{equation}
Consequently,
\begin{equation}
	(p_j-q_j)-(p_a-q_a)
	=(p_j-p_a)+(q_a-q_j)\ge m_q.
	\label{eq:coordinate_difference}
\end{equation}
For arbitrary real numbers $u,v$, $u^2+v^2\ge (v-u)^2/2$. Applying this inequality to $u=p_a-q_a$ and $v=p_j-q_j$, and using Eq.~\eqref{eq:coordinate_difference}, gives
\begin{align}
	\lVert p-q\rVert_2^2
	&\ge (p_a-q_a)^2+(p_j-q_j)^2\\
	&\ge \frac{m_q^2}{2}.
\end{align}
Thus, whenever $m_q>0$ and $h_p\neq h_q$,
\begin{equation}
	\frac{2\lVert p-q\rVert_2^2}{m_q^2}\ge1,
\end{equation}
so $\Psi(p,q;x)=1$. If $m_q=0$, $\Psi$ equals one by definition. This proves Eq.~\eqref{eq:pointwise_certificate}. \hfill$\square$

Taking the expectation of Eq.~\eqref{eq:pointwise_certificate} over $Q_X$ immediately yields
\begin{equation}
	d_Q(h_p,h_q):=\Pr_{Q_X}[h_p(x)\neq h_q(x)]
	\le \mathcal C_Q(p,q).
	\label{eq:population_disagreement}
\end{equation}

\subsubsection{Target-risk sandwich}

\noindent\textbf{Lemma 2.}
For any two classifiers $h_p,h_q$ under the multiclass $0$--$1$ loss,
\begin{equation}
	\big|R_Q(h_p)-R_Q(h_q)\big|\le d_Q(h_p,h_q).
	\label{eq:risk_difference}
\end{equation}

\noindent\emph{Proof.}
For every $(x,y)$,
\begin{equation}
	\left|
	\mathbf 1[h_p(x)\neq y]-\mathbf 1[h_q(x)\neq y]
	\right|
	\le \mathbf 1[h_p(x)\neq h_q(x)].
\end{equation}
Taking expectations and applying $|\mathbb E Z|\le\mathbb E|Z|$ proves Eq.~\eqref{eq:risk_difference}. \hfill$\square$

Since $R_Q(h_q)=\eta_Q$, Lemma~2 and Eq.~\eqref{eq:population_disagreement} imply
\begin{equation}
	\eta_Q-\mathcal C_Q(p,q)
	\le R_Q(h_p)\le
	\eta_Q+\mathcal C_Q(p,q).
\end{equation}
Intersecting this interval with $[0,1]$ proves Theorem~1. Notice that only an upper bound on $d_Q(h_p,h_q)$ is available. Therefore, the generally invalid expression $|\eta_Q-\mathcal C_Q|$ must not be used as a lower bound.

\subsubsection{Teacher-risk uncertainty interval}

If external calibration, an independently labeled diagnostic set, or an explicitly stated assumption provides
\begin{equation}
	\underline\eta\le\eta_Q\le\overline\eta,
\end{equation}
then Theorem~1 implies the robust interval
\begin{equation}
	\max\{0,\underline\eta-\mathcal C_Q(p,q)\}
	\le R_Q(h_p)\le
	\min\{1,\overline\eta+\mathcal C_Q(p,q)\}.
	\label{eq:robust_teacher_interval}
\end{equation}
The interval $[\underline\eta,\overline\eta]$ is not identifiable from $Q_X$ alone: the same unlabeled marginal may be paired with different target labeling functions. Accordingly, Eq.~\eqref{eq:robust_teacher_interval} must not be described as fully observable in a label-free training protocol. In our analysis, target labels may be used only for post-hoc oracle evaluation of $\eta_Q$, never for adaptation or model selection.

\subsubsection{Finite-sample certificate}

Let $S_U=\{x_i\}_{i=1}^n$ be an i.i.d. unlabeled sample from $Q_X$ that is independent of all training, hyperparameter selection, and stopping decisions used to obtain $p$ and $q$. Define
\begin{equation}
	\widehat{\mathcal C}_{S_U}(p,q)
	=\frac{1}{n}\sum_{i=1}^{n}\Psi(p,q;x_i).
\end{equation}
Because $0\le\Psi\le1$, Hoeffding's inequality gives, with probability at least $1-\delta$,
\begin{equation}
	\mathcal C_Q(p,q)
	\le \widehat{\mathcal C}_{S_U}(p,q)
	+\sqrt{\frac{\ln(1/\delta)}{2n}}.
	\label{eq:finite_certificate}
\end{equation}
Combining Eqs.~\eqref{eq:robust_teacher_interval} and \eqref{eq:finite_certificate} yields
\begin{equation}
	R_Q(h_p)\le
	\min\!\left\{1,
	\overline\eta+
	\widehat{\mathcal C}_{S_U}(p,q)
	+\sqrt{\frac{\ln(1/\delta)}{2n}}
	\right\}.
	\label{eq:finite_risk_bound}
\end{equation}
If the same transductive target sample is used both to train the models and to compute $\widehat{\mathcal C}$, the independence condition is violated and Eq.~\eqref{eq:finite_certificate} does not follow from Hoeffding's inequality. In that case, $\widehat{\mathcal C}$ is reported only as an empirical diagnostic unless cross-fitting, a uniform-convergence argument, or a PAC-Bayes analysis is introduced.

\subsubsection{Relation to Dynamic CMKD}

Let $p=P_{\mathrm{stu}}$, $q=P_{\mathrm{tea}}^*$, $c=\exp[-D_{\mathrm{KL}}(p\|q)]$, and $G(r)=1-\lVert r\rVert_2^2$. Since
\begin{equation}
	G\!\left(\frac{p+q}{2}\right)
	=\frac{G(p)+G(q)}{2}+\frac14\lVert p-q\rVert_2^2,
\end{equation}
Eq. (9) of the main paper can be rewritten exactly as
\begin{align}
	\mathcal L_{\mathrm{cmkd}}^*
	=&\frac{\alpha(1+c)}{2}G(p)
	+\left[\frac{\alpha(1-c)}{2}+\beta\right]G(q)\nonumber\\
	&+\frac{\alpha(1-c)}{4}\lVert p-q\rVert_2^2.
	\label{eq:cmkd_decomposition}
\end{align}
Thus, Dynamic CMKD contains an explicit consistency penalty whose coefficient increases monotonically with $D_{\mathrm{KL}}(p\|q)$. This establishes alignment between the training objective and the distribution-discrepancy component of Eq.~\eqref{eq:certificate}. It does not, by itself, prove monotonic decrease of the unweighted quantity $\mathbb E\lVert p-q\rVert_2^2$, because the coefficient and the other Gini terms co-evolve during optimization.

Likewise, Theorem~1 does not prove that the class-anchor stream reduces $\eta_Q$. A high teacher margin represents low predictive ambiguity, not semantic correctness. Any claim that class anchors improve teacher correctness must be supported empirically, for example by oracle teacher pseudo-label accuracy that is used only for analysis.

\begin{comment}
	\subsubsection{Proof of the DSSG--PAC corollary}
	
	Applying Lemma~2 directly to the final DSSG classifier $h_D$ and final DSSG-PAC classifier $h_P$ gives
	\begin{equation}
		\big|R_Q(h_P)-R_Q(h_D)\big|
		\le d_Q(h_P,h_D),
	\end{equation}
	which proves Corollary~1. This statement compares the final deployed classifiers and is therefore different from comparing online and cached teachers inside a single PAC trajectory. It certifies the maximum possible accuracy gap from final-model disagreement, but it does not determine which model is more accurate on the disagreement set.
	
	\subsubsection{Scope of the guarantees}
	
	The analysis proves that: (i) teacher--student disagreement is controlled by a margin-normalized distribution discrepancy; (ii) the student risk lies in a certified neighborhood of the teacher risk; and (iii) final DSSG--PAC/DSSG disagreement bounds their absolute target-risk gap. It does not prove that a larger margin implies a more accurate teacher, that class anchors necessarily reduce teacher error, that Dynamic CMKD monotonically decreases an unweighted consistency loss, or that either DSSG or DSSG-PAC universally improves target accuracy without assumptions on the unknown target labeling function.
\end{comment}

\clearpage
\section{Supplementary Material}

\subsection{Optimization of Baseline Hyper-parameters $\lambda_1$ and $\lambda_2$}
\label{appendix:lambda12}  
\begin{figure*}[t]
	\centering
	\includegraphics[width=1.0\linewidth]{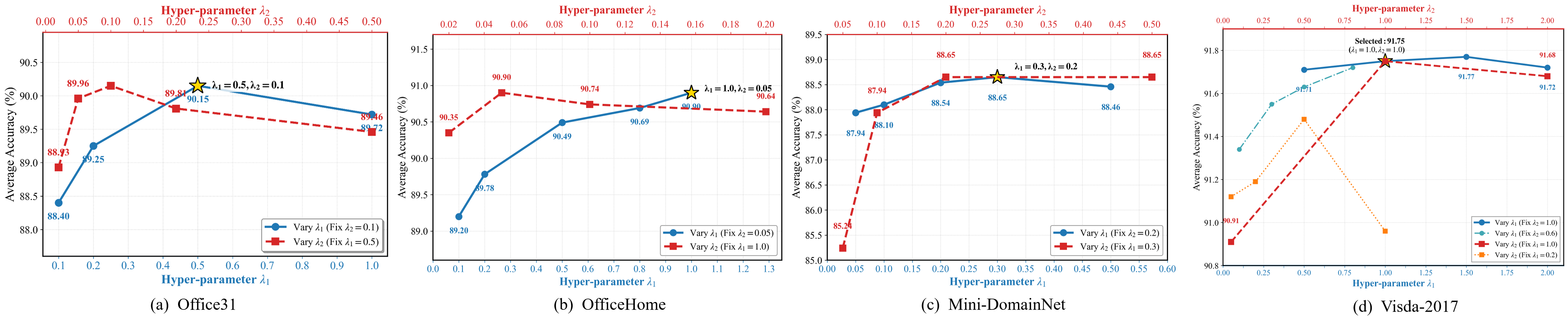}
	\caption{Sensitivity of the ImCapFusionSFDA baseline to $\lambda_1$ (blue solid) and $\lambda_2$ (red dashed). The star ($\star$) indicates the selected optimal configuration.}
	\label{fig:baseline_sensitivity}
\end{figure*}

Although ImCapSFDA reports benchmark results, its optimal $\lambda_1$ and $\lambda_2$ settings are not provided. We therefore optimize these hyperparameters for the reproduced baseline, as shown in Fig.~\ref{fig:baseline_sensitivity}, and directly inherit the selected settings for DSSG without additional tuning.

\begin{figure*}[t]    
	\centering
	\begin{adjustbox}{width=0.95\linewidth,center}
		\includegraphics{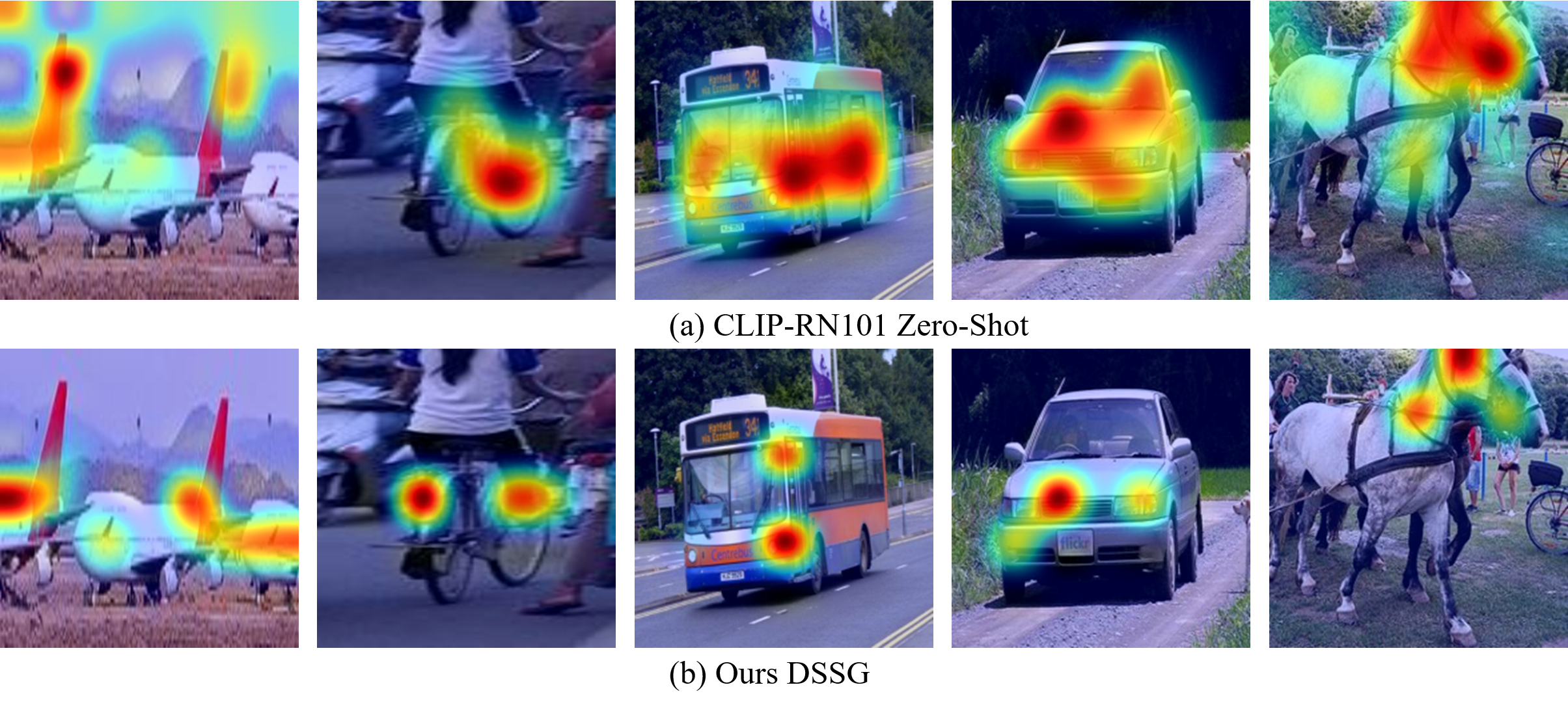}
	\end{adjustbox} 
	\caption{Grad-CAM visualization samples.}
	\label{Fig-gradcam-vis} 
\end{figure*}

\subsection{Activation Map Visualization}
\label{sec:appendix_gradcam}
As shown in Fig.~\ref{Fig-gradcam-vis}, DSSG exhibits more concentrated activations on discriminative object regions and reduced responses to background regions compared with the baseline, indicating improved task-relevant semantic focus.

\end{document}